\documentclass[letterpaper]{article} 
\usepackage{aaai23}  
\usepackage{times}  
\usepackage{helvet}  
\usepackage{courier}  
\usepackage[hyphens]{url}  
\usepackage{graphicx} 
\usepackage{natbib}  
\usepackage{caption} 
\usepackage{algorithm}
\usepackage{algorithmic}
\usepackage{amsmath, fixmath, dsfont}
\usepackage{amssymb}
\usepackage{esvect}
\usepackage{array}
\usepackage{multirow}
\usepackage{textcomp}
\usepackage[svgnames]{xcolor}
\usepackage{booktabs}
\newcommand{\ie}{\textit{i}.\textit{e}.}
\newcommand{\eg}{\textit{e}.\textit{g}.}
\usepackage{newfloat}
\usepackage{listings}
\DeclareCaptionStyle{ruled}{labelfont=normalfont,labelsep=colon,strut=off} 
\floatstyle{ruled}
\newfloat{listing}{tb}{lst}{}
\floatname{listing}{Listing}
\title{Rethinking Rotation Invariance with Point Cloud Registration}
\author {
    Jianhui Yu,
    Chaoyi Zhang,
    Weidong Cai
}
\affiliations {
    School of Computer Science, University of Sydney, Australia\\
    \{jianhui.yu, chaoyi.zhang, tom.cai\}@sydney.edu.au
}

\begin{document}

\maketitle

\begin{abstract}
Recent investigations on rotation invariance for 3D point clouds have been devoted to devising rotation-invariant feature descriptors or learning canonical spaces where objects are semantically aligned.
Examinations of learning frameworks for invariance have seldom been looked into.
In this work, we review rotation invariance in terms of point cloud registration and propose an effective framework for rotation invariance learning via three sequential stages, namely rotation-invariant shape encoding, aligned feature integration, and deep feature registration.
We first encode shape descriptors constructed with respect to reference frames defined over different scales, e.g., local patches and global topology, to generate rotation-invariant latent shape codes.
Within the integration stage, we propose Aligned Integration Transformer to produce a discriminative feature representation by integrating point-wise self- and cross-relations established within the shape codes.
Meanwhile, we adopt rigid transformations between reference frames to align the shape codes for feature consistency across different scales.
Finally, the deep integrated feature is registered to both rotation-invariant shape codes to maximize feature similarities, such that rotation invariance of the integrated feature is preserved and shared semantic information is implicitly extracted from shape codes.
Experimental results on 3D shape classification, part segmentation, and retrieval tasks prove the feasibility of our work.
Our project page is released at: https://rotation3d.github.io/.
\end{abstract}

\section{Introduction}
Point cloud analysis has recently drawn much interest from researchers.
As a common form of 3D representation, the growing presence of point cloud data is encouraging the development of many deep learning methods \citep{qi2017pointnet, pct, SGGpoint}, showing great success for well-aligned point clouds on different tasks.
However, it is difficult to directly apply 3D models to real data as raw 3D objects are normally captured at different viewing angles, resulting in unaligned data samples, which inevitably impact the deep learning models which are sensitive to rotations.
Therefore, rotation invariance becomes an important research topic in the 3D domain.

\begin{figure}[t]
\centering
\includegraphics[width=\linewidth]{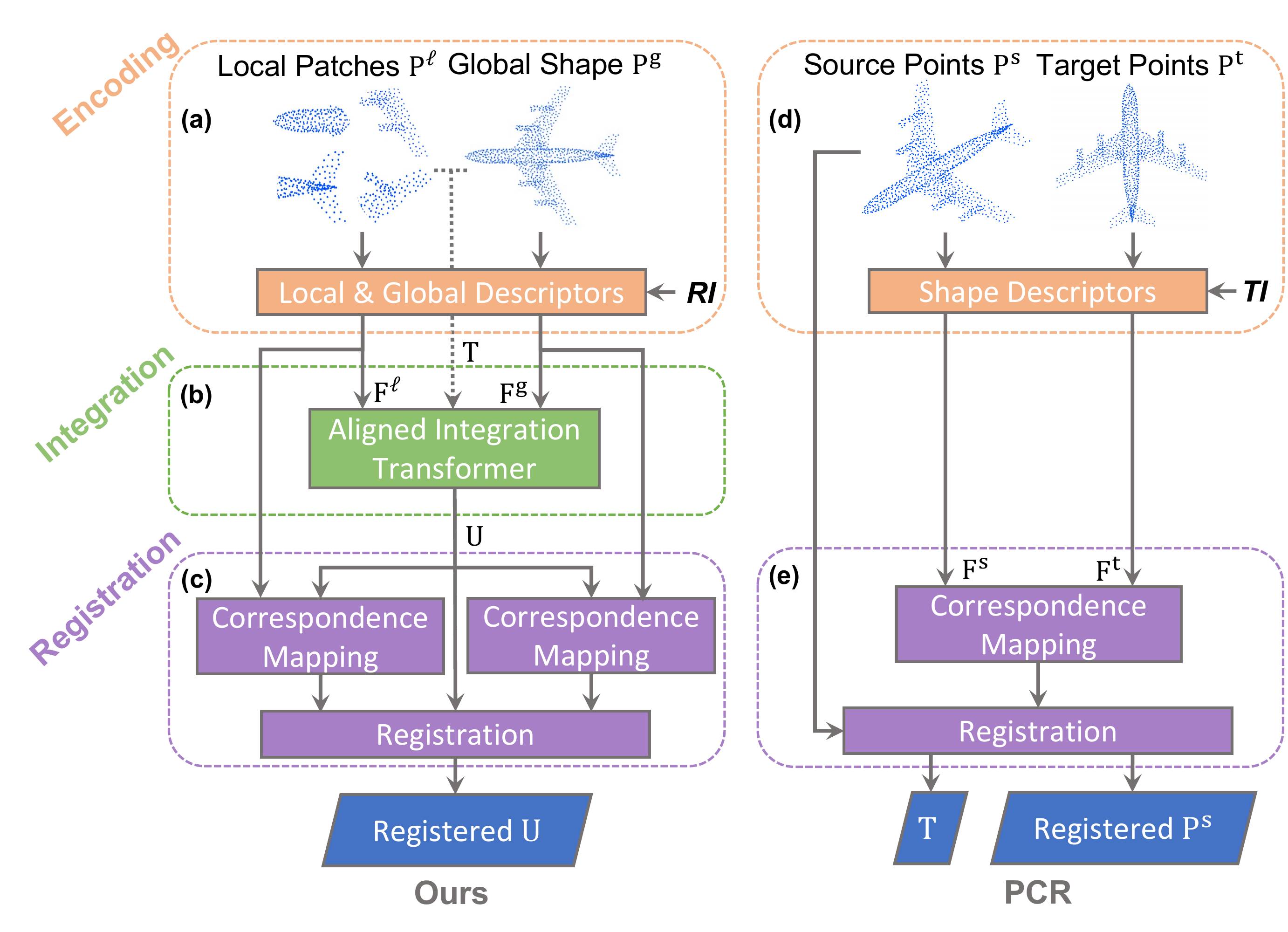}
\caption{
Frameworks of our design (\textbf{left}) and robust point cloud registration (\textbf{right}), where TI and RI are transformation invariance and rotation invariance, and $\mathbf{T}$ is the rigid transformation.
The dotted line indicates the computation of $\mathbf{T}$ between reference frames.
}
\label{fig:pipeline}
\end{figure}

To achieve rotation invariance, a straightforward way is to augment training data with massive rotations which, however, requires a large memory capacity and exhibits limited generalization ability to unseen data~\citep{ri-gcn}.
There are attempts to align 3D inputs to a conical pose \cite{STN, cohen2018spherical}, or to learn rotation robust features via equivariance \cite{vnn, luo2022equivariant}, while these methods are not rigorously rotation-invariant and present noncompetitive performance on 3D shape analysis.
To maintain consistent model behavior under random rotations, some methods \cite{RIconv, clusternet, SGMNet} follow \citet{ppf} to handcraft rotation-invariant point-pair features.
Others \cite{GCA-Conv, li2021closer, LGR-Net} design robust features from equivariant orthonormal bases.
Most of the mentioned works either manipulate model inputs or generate canonical spaces to achieve rotation invariance (RI).
In this work, we review the problem of RI from a different aspect: robust point cloud registration (PCR).
We find that PCR and RI share the \textit{same} goal:
PCR aligns low-dimensional point cloud features (\eg, $xyz$) from the source domain to the target domain regardless of transformations, while RI can be considered to align high-dimensional latent features to rotation-invariant features.
Specifically, the goal of PCR is to explicitly align the source point cloud to the target, both representing the same 3D object, and for RI learning, we implicitly align the final feature representation of a 3D shape to a hidden feature of the same shape, which is universally rotation-invariant to any rotations.


Motivated by this finding, we propose our learning framework in Fig.~\ref{fig:pipeline} with three sequential stages, namely rotation-invariant shape encoding, aligned feature integration, and deep feature registration.
Firstly, we \textbf{(a)} construct and feed point pairs with different scales as model inputs, where we consider local patches $\mathbf{P}^{\ell}$ with small number of points and global shape $\mathbf{P}^{g}$ with the whole 3D points.
Hence, the final feature representation can be enriched by information from different scales.
Low-level rotation-invariant descriptors are thus built on reference frames and encoded to generate latent shape codes $\mathbf{F}^{\ell}$ and $\mathbf{F}^{g}$ following recent PCR work \cite{pan2021robust}.
Secondly, we \textbf{(b)} introduce a variant of transformer~\cite{vaswani2017attention}, Aligned Integration Transformer (AIT), to implicitly integrate information from both self- and cross-attention branches for effective feature integration.
In this way, information encoded from different point scales is aggregated to represent the same 3D object.
Moreover, we consider $\mathbf{F}^{\ell}$ and $\mathbf{F}^{g}$ as \textit{unaligned} since they are encoded from \textit{unaligned} reference frames.
To address the problem, we follow the evaluation technique proposed in PCR~\cite{pan2021robust}, where we use relative rotation information (\textbf{T}) with learnable layers to align $\mathbf{F}^{\ell}$ and $\mathbf{F}^{g}$ for feature consistency.
Finally, to ensure RI of the integrated feature $\mathbf{U}$, we follow PCR to \textbf{(c)} examine the correspondence map of ($\mathbf{F}^{g}$, $\mathbf{U}$) and ($\mathbf{F}^{\ell}$, $\mathbf{U}$), such that the mutual information between a local patch of a 3D object and the whole 3D object is maximized, and RI is further ensured in the final geometric feature.

The contributions of our work are summarized as following three folds:
(1) To our knowledge, we are the first in developing a PCR-cored representation learning framework towards effective RI studies on 3D point clouds.
(2) We introduce Aligned Integration Transformer (AIT), a transformer-based architecture to conduct aligned feature integration for a comprehensive geometry study from both local and global scales.
(3) We propose a registration loss to maintain rotation invariance and discover semantic knowledge shared in different parts of the input object.
Moreover, the feasibility of our proposed framework is successfully demonstrated on various 3D tasks.

\section{Related Work} \label{sec:related_work}
\paragraph{Rotation Robust Feature Learning.}
Networks that are robust to rotations can be equivariant to rotations.
\citet{sphericalcnn} and \citet{cohen2018spherical} project 3D data into a spherical space for rotation equivariance and perform convolutions in terms of spherical harmonic bases.
Some~\cite{spezialetti2020learning, sun2021canonical} learn canonical spaces to unify the pose of point clouds.
Recent works~\cite{luo2022equivariant, vnn, jing2020learning} vectorize the scalar activations and mapping SO(3) actions to a latent space for easy manipulations.
Although these works present competitive results, they cannot be strictly rotation-invariant.
Another way for rotation robustness is to learn rotation-invariant features.
Handcraft point-pair features are rotation-invariant~\cite{RIconv, clusternet, SGMNet}, but they focus on local domains and ignore the global overview of 3D objects.
Others use rotation-equivariant local reference frames (LRFs) \cite{GCA-Conv, thomas2020rotation, ri-gcn} or global reference frames (GRFs) \cite{li2021closer} as model inputs based on principal component analysis (PCA).
However, they may produce inconsistent features across different reference frames, which would limit the representational power.
In contrast to abovementioned methods with rotation robust model inputs or modules, we examine the relation between RI and PCR and propose an effective framework.

\paragraph{3D Robust Point Cloud Registration.}
Given a pair of LiDAR scans, 3D PCR requires an optimal rigid transformation to best align the two scans.
Despite the recent emerging of ICP-based methods \cite{ICP, prnet}, we follow robust correspondence-based approaches in our work~\cite{ppfnet, Deepgmr, geotrans, pan2021robust}, where RI is widely used to mitigate the impact of geometric transformations during feature learning.
Specifically, both \citet{pan2021robust} and \citet{geotrans} analyze the encoding of transformation-robust information and introduce a rotation-invariant module with contextual information into their registration pipeline.
All these methods showing impressive results are closely related to rotation invariance.
We hypothesize that the learning framework of RI can be similar to PCR, and we further prove in experiments that our network is feasible and able to achieve competitive performance on rotated point clouds.

\paragraph{Transformers in 3D Point Clouds.}
Transformers \cite{img16, swin} applied to 2D vision have shown great success, and they are gaining prominence in 3D point clouds.
For example, \citet{pt} uses vectorized self-attention \cite{vaswani2017attention} and positional embedding for 3D modeling.
\citet{pct} proposes offset attention for noise-robust geometric representation learning.
Cross-attention is widely employed for semantic information exchange \cite{geotrans, yu2021cofinet}, where feature relations between the source and target domains are explored.
Taking advantage of both, we design a simple yet effective feature integration module with self and cross relations.
In addition, transformation-related embeddings are introduced for consistent feature learning.

\paragraph{Contrastive Learning with 3D Visual Correspondence.}
Based on visual correspondence, contrastive learning aims to train an embedding space where positive samples are pushed together whereas negative samples are separated away~\cite{he2020momentum}.
The definition of positivity and negativity follows the visual correspondence maps, where pairs with high confidence scores are positive otherwise negative.
Visual correspondence is important in 3D tasks, where semantic information extracted from matched point pairs improves the network's understanding on 3D geometric structures.
For example, PointContrast \cite{pointcontrast} explores feature correspondence across multiple views of one 3D point cloud with InfoNCE loss \cite{CPC}, increasing the model performance for downstream tasks.
Info3D \cite{info3d} and CrossPoint \cite{crosspoint} minimize the semantic difference of point features under different poses.
We follow the same idea by registering the deep features to rotation-invariant features at intermediate levels, increasing feature similarities in the embedding space to ensure rotation invariance.

\section{Method} \label{sec:method}
Given a 3D point cloud including $N_{in}$ points with $xyz$ coordinates $\mathbf{P}=\{p_{i} \in \mathbb{R}^{3}\}_{i=1}^{N_{in}}$, we aim to learn a shape encoder $f$ that is invariant to 3D rotations: $f(\mathbf{P}) = f(\mathbf{R}\mathbf{P})$, where $\mathbf{R} \in SO(3)$ and SO(3) is the rotation group.
RI can be investigated and achieved through three stages, namely rotation-invariant shape encoding (Section~\ref{sec:rf}), aligned feature integration (Section~\ref{sec:transformer}), and deep feature registration (Section~\ref{sec:dfr}).

\subsection{Rotation-Invariant Shape Encoding} \label{sec:rf}
In this section, we first construct the input point pairs from local and global scales based on reference frames, following the idea of~\citet{pan2021robust} to obtain low-level rotation-invariant shape descriptors from LRFs and GRF directly. Then we obtain latent shape codes via two set abstraction layers as in PointNet++~\cite{qi2017pointnet2}.

\paragraph{Rotation Invariance for Local Patches.}
To construct rotation-invariant features on LRFs, we hope to construct an orthonormal basis for each LRF as $p \in \mathbb{R}^{3 \times 3}$.
Given a point $p_i$ and its neighbor $p_{j} \in \mathcal{N}(p_{i})$, we choose $\vv{x_{i}}^{\ell}=\vv{p_mp_i}/\|\vv{p_mp_i}\|_{2}$, where $p_{m}$ is the barycenter of the local geometry and $\|\cdot\|_{2}$ is L2-norm.
We then define $\vv{z_{i}}^{\ell}$ following~\citet{stablepca} to have the same direction as an eigenvector, which corresponds to the smallest eigenvalue via eigenvalue decomposition (EVD):
\begin{equation} \label{eq:covariance}
\small
    \mathbf{\Sigma}^{\ell}_{i} = \sum_{j=1}^{|\mathcal{N}(p_i)|} \alpha_{j}\left(\vv{p_ip_j}\right)\left(\vv{p_ip_j}\right)^{\top}, \: \alpha_{j}=\frac{d - \left\|\vv{p_ip_j}\right\|_{2}}{\sum_{j=1}^{|\mathcal{N}(p_i)|} d - \left\|\vv{p_ip_j}\right\|_{2}},
\end{equation}
where $\alpha_{j}$ is a weight parameter, allowing nearby $p_j$ to have large contribution to the covariance matrix, and $d$ is the maximum distance between $p_i$ and $p_j$.
Finally, we define $\vv{y_{i}}^{\ell}$ as $\vv{z_{i}}^{\ell} \times \vv{x_{i}}^{\ell}$.
RI is introduced to $p_{i}$ with respect to its neighbor $p_{j}$ as ${p}_{ij}^{\ell} = \vv{p_ip_j}^{\top}\mathbf{M}^{\ell}_{i}$.
Proofs of the equivariance of $\mathbf{M}^{\ell}_{i}$ and invariance of ${p}_{ij}^{\ell}$ are shown in the supplementary material.
The latent shape code $\mathbf{F}^{\ell} \in \mathbb{R}^{N \times C}$ is obtained via PointNet++ and max-pooling.

\paragraph{Rotation Invariance for Global Shape.}
We apply PCA as a practical tool to obtain RI in a global scale.
Similar to Eq.~\ref{eq:covariance}, PCA is performed by $\frac{1}{N_{0}}\sum_{i=1}^{N_{0}}(\vv{p_mp_i})(\vv{p_mp_i})^{\top} = \mathbf{U}^{g}\mathbf{\Lambda}^{g}{\mathbf{U}^{g}}^{\top}$, where $p_m$ is the barycenter of $\mathbf{P}$, $\mathbf{U}^{g}=[\vv{u_{1}}^{g}, \vv{u_{2}}^{g}, \vv{u_{3}}^{g}]$ and $\mathbf{\Lambda}^{g}=\operatorname{diag}(\lambda^{g}_{1},\lambda^{g}_{2},\lambda^{g}_{3})$ are eigenvector and eigenvalue matrices.
We take $\mathbf{U}^{g}$ as the orthonormal basis $\mathbf{M}^{g}=[\vv{x}^{g}, \vv{y}^{g}, \vv{z}^{g}]$ for GRF.
By transforming point $p_i$ with $\mathbf{U}^{g}$, the shape pose is canonicalized as $p_i^{g} = p_i\mathbf{M}^{g}$.
Proof of the RI of $p_i^{g}$ is omitted for its simplicity, and $\mathbf{F}^{g} \in \mathbb{R}^{N \times C}$ is obtained following PointNet++.

\paragraph{Sign Ambiguity.}
EVD introduces sign ambiguity for eigenvectors, which negatively impacts the model performance \cite{pca_issue}.
The description of sign ambiguity states that for a random eigenvector $\vv{u}$, $\vv{u}$ and $\vv{u}^{\prime}$, with $\vv{u}^{\prime}$ having an opposite direction to $\vv{u}$, are both acceptable solutions to EVD.
To tackle this issue, we simply force $\vv{z_{i}}^{\ell}$ of LRF to follow the direction of $\vv{op_i}$, with $o$ being the origin of the world coordinate.
We disambiguate basis vectors in $\mathbf{M}^{g}$ by computing an inner product with $\vv{p_mp_i}, \forall i \in N_{0}$.
Taking $\vv{x}^{g}$ for example, its direction is conditioned on the following term:
\begin{equation} \label{eq:direction}
\small
    \vv{x}^{g} = 
    \begin{cases}
         \vv{x}^{g}, & \text{if } S_{x} \geq \frac{N_{0}}{2} \\
         \vv{x}^{\prime g},& \text{otherwise}
    \end{cases}, \quad S_{x} = \sum_{i=1}^{N_{0}} \mathds{1}[\langle \vv{x}^{g}, \vv{p_mp_i}\rangle],
\end{equation}
where $\langle \cdot,\cdot \rangle$ is the inner product, $\mathds{1}[\cdot]$ is a binary indicator that returns 1 if the input argument is positive, otherwise 0.
$S_{x}$ denotes the number of points where $\vv{x}^{g}$ and $\vv{p_mp_i}$ point to the same direction.
The same rule is applied to disambiguate $\vv{y}^{g}$ and $\vv{z}^{g}$ by $S_{y}$ and $S_{z}$.
Besides, as mentioned in \citet{li2021closer}, $\mathbf{M}^{g}$ might be non-rotational (e.g., reflection).
To ensure $\mathbf{M}^{g}$ a valid rotation, we simply reverse the direction of the basis vector whose $S$ value is the smallest.
More analyses on sign ambiguity are in the supplementary material.

\subsection{Aligned Feature Integration} \label{sec:transformer}
Transformer has been widely used in 3D domain to capture long-range dependencies~\cite{3dmedpt}.
In this section, we introduce Aligned Integration Transformer (AIT), an effective transformer to align latent shape codes with relative rotation angles and integrate information via attention-based integration~\cite{fusenet}.
Within each AIT module, we first apply Intra-frame Aligned Self-attention on $\mathbf{F}^{\ell}$ and we do not encode $\mathbf{F}^{g}$, which is treated as supplementary information to assist local geometry learning with the global shape overview.
We discuss that encoding $\mathbf{F}^{g}$ via self-attention can increase model overfitting, thus lowering the model performance.
We will validate our discussion in Section~\ref{sec:ablation}.
Inter-frame Aligned Cross-attention is applied on both $\mathbf{F}^{\ell}$ and $\mathbf{F}^{g}$, and we use Attention-based Feature Integration module for information Aggregation.


\paragraph{Preliminary: Offset Attention.}
AIT utilizes offset attention~\cite{pct} for noise robustness.
In the following, we use subscripts \textit{sa} and \textit{ca} to denote implementations related to self- and cross-attention, respectively.
We first review offset attention as follows:
\begin{equation} \label{eq:offset}
\small
    \begin{gathered}
        \mathbf{F} = \boldsymbol{\phi}(\mathbf{F}_{oa}) + \mathbf{F}_{in}, \; \mathbf{F}_{oa} = \mathbf{F}_{in} - \|\text{softmax}(\mathbf{A})\|_{1}\mathbf{v}, \;
        \mathbf{A} = \mathbf{q}\mathbf{k}^{\top},
    \end{gathered}
\end{equation}
where $\mathbf{q}=\mathbf{F}_{in}\mathbf{W}_{q}$, $\mathbf{k}=\mathbf{F}_{in}\mathbf{W}_{k} \in \mathbb{R}^{N \times d}$, and $\mathbf{v}=\mathbf{F}_{in}\mathbf{W}_{v} \in \mathbb{R}^{N \times C}$ are query, key, and value embeddings, and $\mathbf{W}_{q}$, $\mathbf{W}_{k} \in \mathbb{R}^{C \times d}$, $\mathbf{W}_{v} \in \mathbb{R}^{C \times C}$ are the corresponding projection matrices.
$\|\cdot\|_{1}$ is L1-norm and $\boldsymbol{\phi}$ denotes a multi-layer perceptron (MLP).
$\mathbf{F}_{oa}$ is offset attention-related feature and $\mathbf{A}\in \mathbb{R}^{N \times N}$ is the attention logits.

\begin{figure*}[t]
    \centering
    \includegraphics[width=0.85\linewidth]{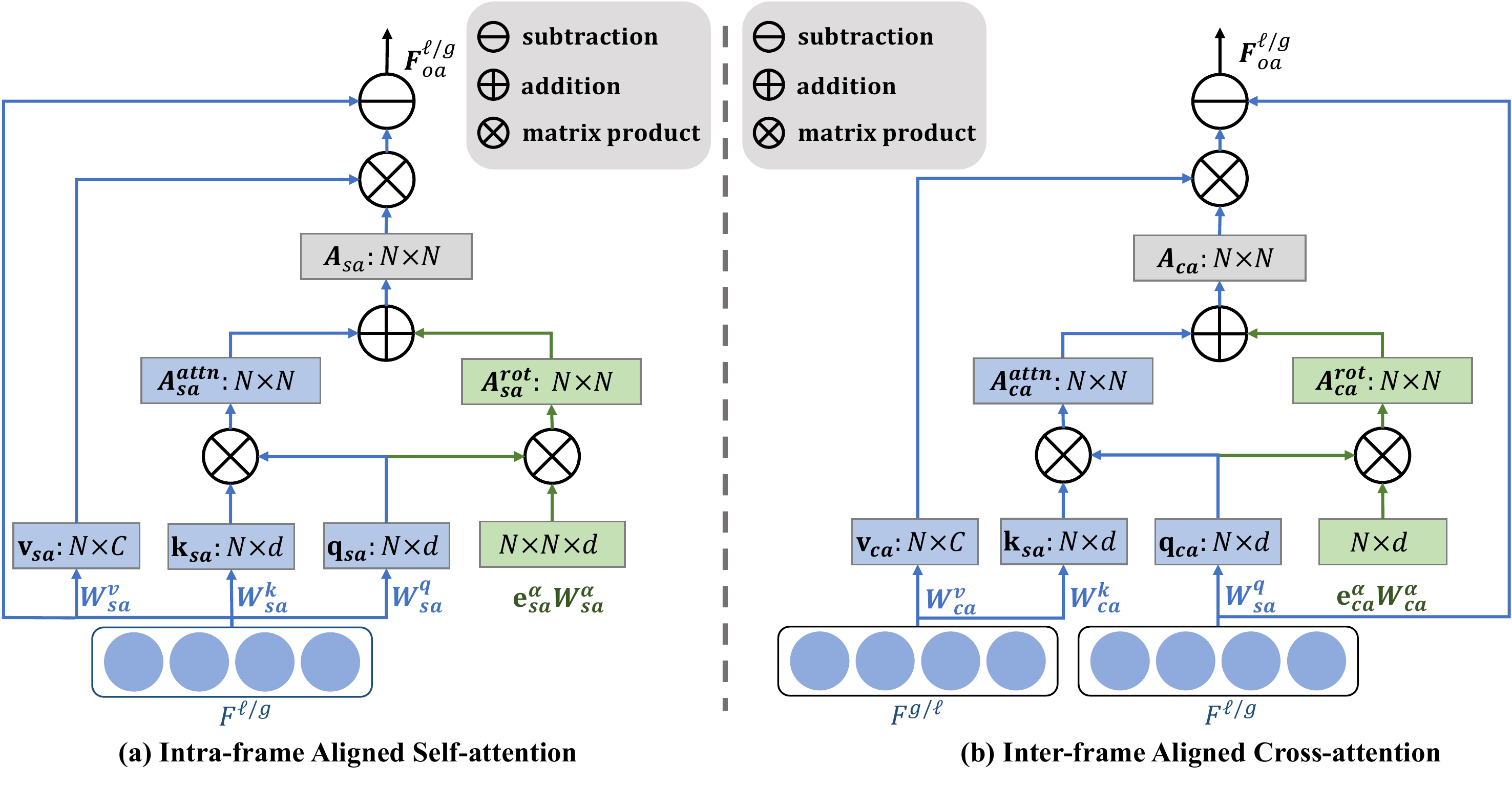}
    \caption{Illustrations of (a) Intra-frame Aligned Self-attention and (b) Inter-frame Aligned Cross-attention modules. Note that we only present processes for computing $\mathbf{F}_{oa}$ in both modules.}
    \label{fig:attn}
\end{figure*}

\paragraph{Intra-frame Aligned Self-attention.}
Point-wise features of $\mathbf{F}^{\ell}$ are encoded from \textit{unaligned} LRFs, so direct implementation of self-attention on $\mathbf{F}^{\ell}$ can cause feature inconsistency during integration.
To solve this problem, rigid transformations between distinct LRFs are considered, which are explicitly encoded and injected into point-wise relation learning process.
We begin by understanding the transformation between two LRFs.
For any pair of local orthonormal bases $\mathbf{M}^{\ell}_{i}$ and $\mathbf{M}^{\ell}_{j}$, a rotation can be easily derived $\mathrm{\Delta}\mathbf{R}_{ji} = \mathbf{M}^{\ell}_{i} {\mathbf{M}^{\ell}_{j}}^{\top}$ and translation is defined as $\mathrm{\Delta}\mathbf{t}_{ji} = o^{\ell}_{i} - o^{\ell}_{j}$, where $o^{\ell}_{i/j}$ indicates the origin.
In our work, the translation part is intentionally ignored, where we show in the supplementary material that by keeping both rotation and translation information, the model performance decreases.

Although $\mathrm{\Delta}\mathbf{R}_{ji}$ is invariant to rotations, we do not directly project it into the embedding space, as it is sensitive to the order of matrix product: $\mathrm{\Delta}\mathbf{R}_{ji} \neq \mathrm{\Delta}\mathbf{R}_{ij}$, giving inconsistent rotation information when the product order is not maintained.
To address this issue, we construct our embedding via the relative rotation angle $\mathrm{\Delta} \alpha_{ji}$ between $\mathbf{M}^{\ell}_{i}$ and $\mathbf{M}^{\ell}_{j}$, which is normally used in most PCR works~\cite{rpmnet, pan2021robust} for evaluations.
The relative rotation angle $\mathrm{\Delta} \alpha_{ji}$ is computed as:
\begin{equation} \label{eq:metric}
\centering
\small
    \mathrm{\Delta} \alpha_{ji} = \arccos \left(\frac{\operatorname{Trace} \left(\mathrm{\Delta}\mathbf{R}_{ji}\right) - 1}{2}\right) \frac{180}{\pi} \in [0, \pi],
\end{equation}
where it is easy to see that $\mathrm{\Delta} \alpha_{ji}=\mathrm{\Delta} \alpha_{ij}$.
We further apply sinusoidal functions on $\mathrm{\Delta} \alpha_{ji}$ to generate $N^{2}$ pairs of angular embeddings $\mathbf{e}^{\alpha} \in \mathbb{R}^{N \times N \times d}$ for all $N$ points as:
\begin{equation} \label{eq:pos}
\centering
\small
e_{i,j,2k}^{\alpha}=\sin \left(\frac{\mathrm{\Delta} \alpha_{ji} / t_{\alpha}}{10000^{2k / d}}\right), \;
e_{i,j,2k+1}^{\alpha}=\cos \left(\frac{\mathrm{\Delta} \alpha_{ji} / t_{\alpha}}{10000^{2 k / d}}\right),
\end{equation}
where $t_{\alpha}$ controls the sensitivity to angle variations.

Finally, we inject $\mathbf{e}^{\alpha}$ into offset attention and learn intra-frame aligned feature $\mathbf{F}^{\ell}_{\text{IAS}}$ via self-attention as follows:
\begin{equation} \label{eq:intra}
\small
\centering
\begin{gathered}
    \mathbf{F}^{\ell}_{\text{IAS}} = \boldsymbol{\phi}\left(\mathbf{F}_{oa}^{\ell}\right) + \mathbf{F}^{\ell}, 
    \mathbf{F}_{oa}^{\ell} = \mathbf{F}^{\ell} - \|\operatorname{softmax}(\mathbf{A}_{sa})\|_{1}\mathbf{v}_{sa}, \\
    \mathbf{A}_{sa} = \mathbf{A}_{sa}^{attn} + \mathbf{A}_{sa}^{rot}, 
    \mathbf{A}_{sa}^{attn} = \mathbf{q}_{sa}\mathbf{k}_{sa}^{\top},
    \mathbf{A}_{sa}^{rot} = \mathbf{q}_{sa}
    (\mathbf{e}_{sa}^{\alpha}\mathbf{W}_{sa}^{\alpha})^{\top},
\end{gathered}
\end{equation}
where $\mathbf{q}_{sa}/\mathbf{k}_{sa}/\mathbf{v}_{sa}=\mathbf{F}^{\ell}\mathbf{W}^{\mathbf{q}}_{sa}/\mathbf{F}^{l}\mathbf{W}^{\mathbf{k}}_{sa}/\mathbf{F}^{l}\mathbf{W}^{\mathbf{v}}_{sa}$, $\mathbf{W}_{sa}^{\alpha} \in \mathbb{R}^{d \times d}$ is a linear projection to refine the learning of $\mathbf{e}_{sa}^{\alpha}$, and $\mathbf{A}_{sa}$ is the attention logits.
The same process can be performed for $\mathbf{F}^{g}$ by swapping the index $\ell$ and $g$.
Detailed illustrations are shown in Fig.~\ref{fig:attn}~(a).

\paragraph{Inter-frame Aligned Cross-attention.}
Semantic information exchange between $\mathbf{F}^{\ell}$ and $\mathbf{F}^{g}$ in the feature space is implemented efficiently by cross-attention~ \cite{crossvit}.
Since $\mathbf{F}^{\ell}$ and $\mathbf{F}^{g}$ are learned from different coordinate systems, inter-frame transformations should be considered for cross-consistency between $\mathbf{F}^{\ell}$ and $\mathbf{F}^{g}$.
An illustration of the cross-attention module is shown in Fig.~\ref{fig:attn}~(b).
Computation of inter-frame aligned feature $\mathbf{F}_{\text{IAC}}^{\ell}$ via cross-attention follows a similar way as Eq.~\ref{eq:intra}:
\begin{equation} \label{eq:inter}
\small
\centering
\begin{gathered}
    \mathbf{F}^{\ell}_{\text{IAC}} = \boldsymbol{\phi}\left(\mathbf{F}_{oa}^{\ell}\right) + \mathbf{F}^{\ell}, 
    \mathbf{F}_{oa}^{\ell} = \mathbf{F}^{\ell} - \|\operatorname{softmax}(\mathbf{A}_{ca})\|_{1}\mathbf{v}_{ca}, \\
    \mathbf{A}_{ca} = \mathbf{A}_{ca}^{attn} + \mathbf{A}_{ca}^{rot}, 
    \mathbf{A}_{ca}^{attn} = \mathbf{q}_{ca}\mathbf{k}_{ca}^{\top},
    \mathbf{A}_{ca}^{rot} = \mathbf{q}_{ca}
    (\mathbf{e}_{ca}^{\alpha}\mathbf{W}_{ca}^{\alpha})^{\top},
\end{gathered}
\end{equation}
where $\mathbf{q}_{ca}/\mathbf{k}_{ca}/\mathbf{v}_{ca}=\mathbf{F}^{\ell}\mathbf{W}^{\mathbf{q}}_{ca}/\mathbf{F}^{g}\mathbf{W}^{\mathbf{k}}_{ca}/\mathbf{F}^{g}\mathbf{W}^{\mathbf{v}}_{ca}$.
$\mathbf{A}_{ca}$ is cross-attention logits containing point-wise cross-relations over point features defined across local and global scales.
$\mathbf{e}_{ca}^{\alpha} \in \mathbb{R}^{N \times d}$
is computed via Eq.~\ref{eq:metric} and Eq.~\ref{eq:pos} in terms of the transformation between $\mathbf{M}^{\ell}_{i}$ and $\mathbf{M}^{g}$.
To this end, the geometric features learned between local and global reference frames can be aligned given $\mathbf{e}_{ca}^{\alpha}$, leading to a consistent feature representation.

\paragraph{Attention-based Feature Integration.}
Instead of simply adding the information from both $\mathbf{F}^{\ell}$ and $\mathbf{F}^{g}$, we integrate information by incrementing attention logits.
Specifically, we apply self-attention on $\mathbf{F}^{\ell}$ with attention logits $\mathbf{A}_{sa}$ and cross-attention between $\mathbf{F}^{\ell}$ and $\mathbf{F}^{g}$ with attention logits $\mathbf{A}_{ca}$.
We combine $\mathbf{A}_{sa}$ and $\mathbf{A}_{ca}$ via addition, so that encoded information of all point pairs from a local domain can be enriched by the global context of the whole shape.
Illustration is shown in the supplementary material.
The whole process is formulated as follows:
\begin{equation} \label{eq:fuse}
\small
\begin{gathered}
    \mathbf{U} = \boldsymbol{\phi}\left(\mathbf{F}_{oa}\right) + \mathbf{F}^{\ell}, \\
    \mathbf{F}_{oa} = \mathbf{F}^{\ell} - \|\text{softmax}(\mathbf{A}_{sa} + \mathbf{A}_{ca})\|_{1}(\mathbf{v}_{sa} + \mathbf{v}_{ca}).
\end{gathered}
\end{equation}
Hence, intra-frame point relations can be compensated by inter-frame information communication in a local-to-global manner, which enriches the geometric representations.

\subsection{Deep Feature Registration} \label{sec:dfr}
Correspondence mapping~\cite{dcp, pan2021robust} plays an important role in PCR, and we discuss that it is also critical for achieving RI in our design.
Specifically, although $\mathbf{F}^{\ell}$ and $\mathbf{F}^{g}$ are both rotation-invariant by theory, different point sampling methods and the sign ambiguity will cause the final feature not strictly rotation-invariant.
To solve this issue, we first examine the correspondence map:
\begin{equation} \label{eq:m}
\small
    m\left(\mathcal{X}, \mathcal{Y}\right)= \frac{\operatorname{exp}\left(\Phi_{1}(\mathcal{Y}) \Phi_{2}(\mathcal{X})^{\top}/t\right)}{\sum_{j=1}^{N} \operatorname{exp}\left(\Phi_{1}(\mathcal{Y})\Phi_{2}(\boldsymbol{x}_{j})^{\top}/t\right)},
\end{equation}
where $\Phi_{1}$ and $\Phi_{2}$ are MLPs that project latent embeddings $\mathcal{X}$ and $\mathcal{Y}$ to a shared space, and $t$ controls the variation sensitivity.
It can be seen from Eq.~\ref{eq:m} that the mapping function $m$ reveals feature similarities in the latent space, and it is also an essential part for 3D point-level contrastive learning in PointContrast~\cite{pointcontrast} for the design of InfoNCE losses~\cite{CPC}, which have been proven to be equivalent to maximize the mutual information.
Based on this observation, we propose a registration loss function $\mathcal{L}_{r} = \mathcal{L}^{\ell}_{r} + \mathcal{L}^{g}_{r}$, where $\mathcal{L}^{\ell}_{r}$ and $\mathcal{L}^{g}_{r}$ represent the registration loss of ($\mathbf{F}^{\ell}$,$\mathbf{U}$) and ($\mathbf{F}^{g}$,$\mathbf{U}$).
Mathematically, $\mathcal{L}^{\ell}_{r}$ is defined as follows:
\begin{equation} \label{eq:reg_loss}
\small
    \mathcal{L}^{\ell}_{r} = -\sum_{(i,j) \in M} \operatorname{log}\frac{\operatorname{exp}\left(\Phi_{1}(\mathbf{U}_{j})\Phi_{2}(\mathbf{f}_{i}^{\ell})^{\top}/t\right)}{\sum_{(\cdot, k) \in M} \exp\left(\Phi_{1}(\mathbf{U}_{k})\Phi_{2}(\mathbf{f}_{i}^{\ell})^{\top}/t\right)}.
\end{equation}
The same rule is followed to compute $\mathcal{L}^{g}_{r}$.
Although we follow the core idea of PointContrast, we differ from it in that PointContrast defines positive samples based on feature correspondences computed at the same layer level, while our positive samples are defined across layers.

The intuition for the loss design is that the 3D shape is forced to learn about its local region as it has to distinguish it from other parts of diﬀerent objects. Moreover, we would like to maximize the mutual information between different poses of the 3D shape, as features encoded from different poses should represent the same object, which is very useful in achieving RI in SO(3). 
Moreover, the mutual information between $\mathbf{F}^{\ell}$ and $\mathbf{F}^{g}$ is implicitly maximized, such that shared semantic information about geometric structures can be learned, leading to a more geometrically accurate and discriminative representation.
More details about $\mathcal{L}^{\ell}_{r}$ can be found in the supplementary material.

\section{Experiments} \label{sec:experiments}
We evaluate our model on 3D shape classification, part segmentation, and retrieval tasks under rotations, and extensive experiments are conducted to analyze the network design.
Detailed model architectures for the three tasks are shown in the supplementary material.
Our evaluating protocols are the same as \cite{sphericalcnn}: training and testing the network under azimuthal rotations (z/z); training under azimuthal rotations while testing under arbitrary rotations (z/SO(3)); and training and testing under arbitrary rotations (SO(3)/SO(3)).

\subsection{3D Object Classification}
\paragraph{Synthetic Dataset.}
We first examine the model performance on the synthetic ModelNet40 \cite{wu20153d} dataset.
We sample 1024 points from each data with only $xyz$ coordinates as input features.
Hyper-parameters for training follow the same as \cite{pct}, except that points are downsampled in the order of (1024, 512, 128) with feature dimensions of (3, 128, 256).
We report and compare our model performance with state-of-the-art (SoTA) methods in Table \ref{tab:cls}.
Both rotation sensitive and robust methods achieve great performance under z/z.
However, the former could not generalize well to unseen rotations.
Rotation robust methods like Spherical CNN \cite{sphericalcnn} and SFCNN \cite{sfcnn} achieve competitive results under z/z, but their performance is not consistent on z/SO(3) and SO(3)/SO(3) due to the imperfect projection from points to voxels when using spherical solutions.
We outperform the recent proposed methods \cite{luo2022equivariant, SGMNet, vnn} and achieve an accuracy of 91.0\%, proving the superiority of our framework on classification.

\begin{table}[t]
    \centering
    \resizebox{0.95\columnwidth}{!}{%
    \begin{tabular}{l|ccc|c}
        \toprule
        \textbf{Rotation Sensitive} & z/z  & z/SO(3) & SO(3)/SO(3) & $\mathbf{\Delta}$  \\ 
        \hline
        PointNet \cite{qi2017pointnet}       & 89.2 & 16.2    & 75.5        & 59.3  \\
        PoinNet++ \cite{qi2017pointnet2}     & 89.3 & 28.6    & 85.0     & 56.4  \\
        PCT  \cite{pct}                     & 90.3 & 37.2    & 88.5        & 51.3  \\ 
        \hline
        \textbf{Rotation Robust}          & z/z  & z/SO(3) & SO(3)/SO(3) & $\mathbf{\Delta}$  \\ 
        \hline
        Spherical CNN* \cite{sphericalcnn}  & 88.9 & 76.9    & 86.9        & 10    \\
        SFCNN \cite{sfcnn}                  & \textbf{91.4} & 84.8    & 90.1        & 5.3   \\
        RIConv \cite{RIconv}                & 86.5 & 86.4    & 86.4        & 0.1   \\
        ClusterNet \cite{clusternet}        & 87.1 & 87.1    & 87.1        & 0.0   \\
        PR-InvNet \cite{pr-invnet}          & 89.2 & 89.2    & 89.2        & 0.0   \\
        RI-GCN \cite{ri-gcn}                & 89.5 & 89.5    & 89.5        & 0.0   \\
        GCAConv \cite{GCA-Conv}             & 89.0 & 89.1    & 89.2        & 0.1   \\
        RI-Framework \cite{ri-framework}    & 89.4 & 89.4    & 89.3        & 0.1   \\
        VN-DGCNN \cite{vnn}                 & 89.5 & 89.5    & 90.2        & 0.7   \\
        SGMNet \cite{SGMNet}                & 90.0 & 90.0    & 90.0        & 0.0   \\
        \citet{li2021closer}                & 90.2 & 90.2    & 90.2        & 0.0   \\
        OrientedMP \cite{luo2022equivariant}& 88.4 & 88.4    & 88.9        & 0.5   \\
        ELGANet \cite{ELGA}                 & 90.3 & 90.3    & 90.3        & 0.0   \\
        \hline
        Ours                                & 91.0 & \textbf{91.0}    & \textbf{91.0}        & 0.0   \\
        \bottomrule
    \end{tabular}%
    }
    \caption{
    Classification results on ModelNet40 under rotations.
    * denotes the input type as projected voxels of $2\times 64^{2}$, while the rest take raw points of $1024\times3$ as inputs.
    $\mathbf{\Delta}$ is the absolute difference between z/SO(3) and SO(3)/SO(3).
    } \label{tab:cls}
\end{table}

\begin{table}[t]
    \centering
    \resizebox{0.95\columnwidth}{!}{%
    \begin{tabular}{l|ccc} 
        \toprule
        Method       & z/SO(3) & SO(3)/SO(3) & $\mathbf{\Delta}$  \\ 
        \hline
        PointNet \cite{qi2017pointnet}      & 16.7 & 54.7 & 38.0 \\
        PointNet++ \cite{qi2017pointnet2}   & 15.0 & 47.4 & 32.4 \\
        PCT \cite{pct}                      & 28.5 & 45.8 & 17.3 \\
        \hline
        RIConv \cite{RIconv}                & 78.4 & 78.1 & 0.3 \\
        RI-GCN~\cite{ri-gcn}                & 80.5 & 80.6 & 0.1 \\
        GCAConv \cite{GCA-Conv}             & 80.1 & 80.3 & 0.2 \\
        RI-Framework \cite{ri-framework}    & 79.8 & 79.9 & 0.1 \\
        LGR-Net \cite{LGR-Net}              & 81.2 & 81.4 & 0.2 \\
        VN-DGCNN~\cite{vnn}                 & 79.8 & 80.3 & 0.5 \\
        OrientedMP~\cite{luo2022equivariant}& 76.7 & 77.2 & 0.5 \\
        \hline
        Ours                                & \textbf{86.6} & \textbf{86.3} & 0.3 \\
        \bottomrule
    \end{tabular}%
    }
    \caption{
    Classification results on ScanObjectNN OBJ\_BG under z/SO(3) and SO(3)/SO(3).} 
    \label{tab:scan}
\end{table}

\begin{figure}[t]
\centering
\includegraphics[width=\linewidth]{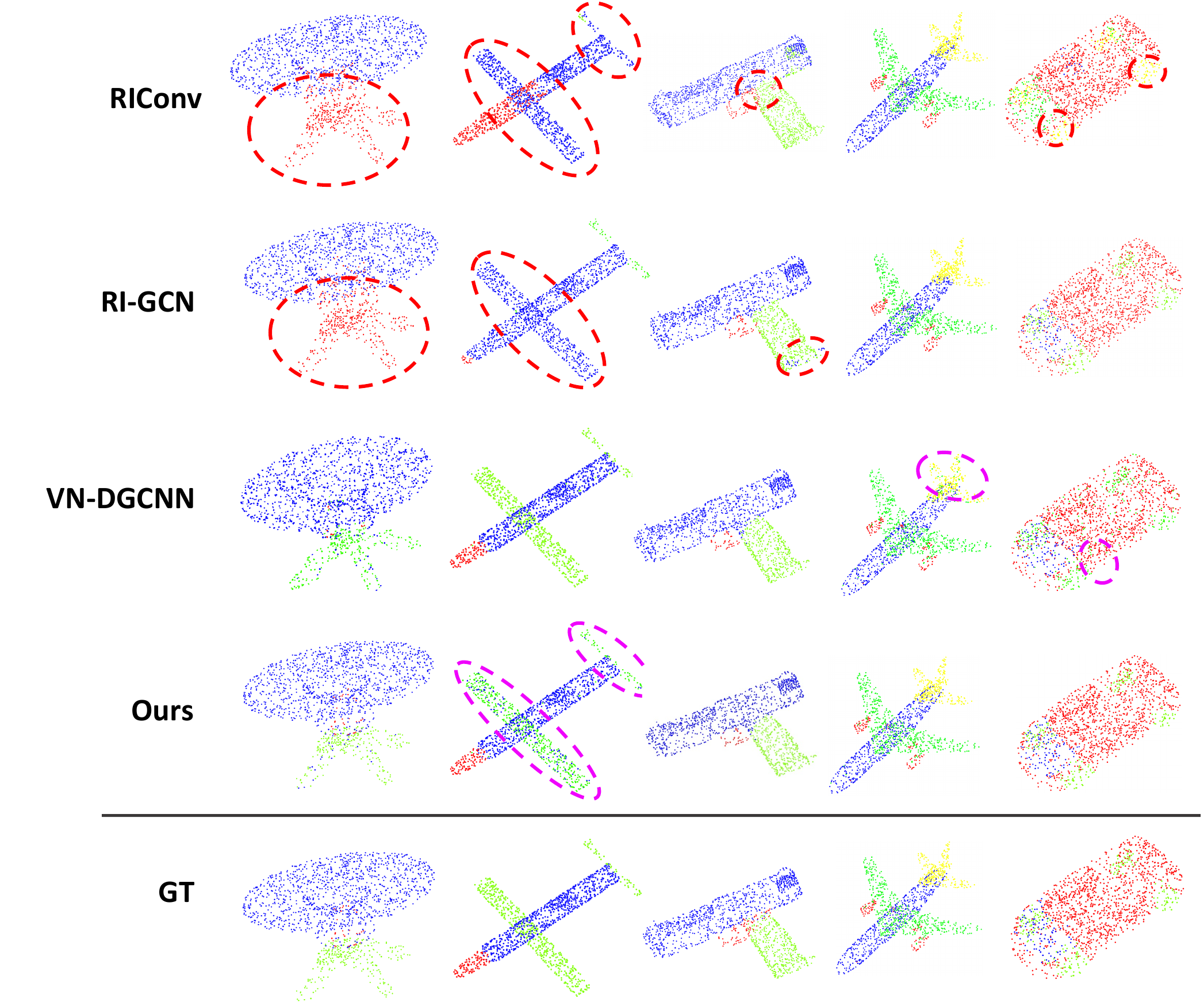}
\caption{
Segmentation comparisons on ShapeNetPart, where ground truth (GT) samples are shown for reference. Red dotted circles indicate obvious failures on certain classes, and purple circles denote the slight difference between our design and VN-DGCNN.
}
\label{fig:seg}
\end{figure}

\begin{table}[t]
    \centering
    \resizebox{0.95\columnwidth}{!}{%
    \begin{tabular}{l|ccc} 
        \toprule
        Method       & z/SO(3) & SO(3)/SO(3) & $\mathbf{\Delta}$  \\ 
        \hline
        PointNet \cite{qi2017pointnet}       & 38.0 & 62.3 & 24.3 \\
        PointNet++ \cite{qi2017pointnet2}    & 48.3 & 76.7 & 28.4 \\
        PCT \cite{pct}                       & 38.5 & 75.2 & 36.7 \\
        \hline
        RIConv \cite{RIconv}                 & 75.3 & 75.5 & 0.2 \\
        RI-GCN \cite{ri-gcn}                 & 77.2 & 77.3 & 0.1 \\
        RI-Framework \cite{ri-framework}     & 79.2 & 79.4 & 0.2 \\
        LGR-Net \cite{LGR-Net}               & 80.0 & 80.1 & 0.1 \\
        VN-DGCNN \cite{vnn}                  & \textbf{81.4} & \textbf{81.4} & 0.0 \\
        OrientedMP \cite{luo2022equivariant} & 80.1 & \underline{80.9} & 0.8 \\
        \hline
        Ours                                 & \underline{80.3} & 80.4 & 0.1 \\
        \bottomrule
    \end{tabular}%
    }
    \caption{Segmentation results on ShapeNetPart under z/SO(3) and SO(3)/SO(3), where the second best results are underlined.} \label{tab:seg}
\end{table}

\paragraph{Real Dataset.}
Experiments are also conducted on a real-scanned dataset.
ScanObjectNN~\cite{scanobjectnn} is a commonly used benchmark to explore the robustness to noisy and deformed 3D objects with non-uniform surface density, which includes 2,902 incomplete point clouds in 15 classes.
We use \textit{OBJ\_BG} subset with the background noise and sample 1,024 points under z/SO(3) and SO(3)/SO(3).
Table~\ref{tab:scan} shows that our model achieves the highest results with excellent consistency with random rotations.

\subsection{3D Part Segmentation}
Shape part segmentation is a more challenging task than object classification.
We use ShapeNetPart \cite{shapenet} for evaluation, where we sample 2048 points with $xyz$ coordinates as model inputs.
The training strategy is the same as the classification task except that the training epoch number is 300.
Part-averaged IoU (mIoU) is reported in Table \ref{tab:seg}, and detailed per-class mIoU values are shown in the supplementary material.
Representative methods such as PointNet++ and PCT are vulnerable to rotations.
Rotation robust methods present competitive results under z/SO(3), where we achieve the second best result of 80.3\%.
We give more details of comparison between VN-DGCNN~\cite{vnn} and our work in the supplementary material, where our method performs better than VN-DGCNN for several classes.
Moreover, qualitative results shown in Fig.~\ref{fig:seg} present that we can achieve visually better results than VN-DGCNN in certain classes such as the airplane and car.
More qualitative results are shown in the supplementary material.

\subsection{3D Shape Retrieval}
We further conduct 3D shape retrieval experiments on ShapeNetCore55 \cite{shapenetcore}, which contains two categories of datasets: normal and perturbed.
We only use the perturbed part to validate our model performance under rotations.
We combine the training and validation sets and validate our method on the testing set following the training policy of \cite{sphericalcnn}.
Experimental results are reported in Table~\ref{tab:shrec}, where the final score is the average value of micro and macro mean average of precision (mAP) as in~\cite{shrec17}.
Similar to the classification task, our method achieves SoTA performance.

\subsection{Ablation Study} \label{sec:ablation}
\begin{table}[t]
    \centering
    \resizebox{0.97\columnwidth}{!}{%
    \begin{tabular}{l|ccc} 
    \toprule
    Method & micro mAP & macro mAP & Score \\ \hline
    Spherical CNN \cite{sphericalcnn} & 0.685     & 0.444     & 0.565 \\
    SFCNN \cite{sfcnn}        & 0.705     & 0.483     & 0.594 \\
    GCAConv \cite{GCA-Conv}      & 0.708     & 0.490     & 0.599 \\
    RI-Framework \cite{ri-framework}  & 0.707     & \textbf{0.510}     & 0.609 \\
    Ours          & \textbf{0.715}     & \textbf{0.510}     & \textbf{0.613} \\
    \bottomrule
    \end{tabular}%
    }
    \caption{Comparisons of SoTA methods on the 3D shape retrieval task.}
    \label{tab:shrec}
\end{table}

\begin{table}
    \centering
    \resizebox{0.9\columnwidth}{!}{%
    \begin{tabular}{l|cc|c|c|cc|c}
    \toprule
    Model & $\mathbf{e}^{\alpha}_{sa}$ & $\mathbf{e}^{\alpha}_{ca}$  & $\mathbf{F}^{g*}$ & $\mathbf{A}_{sa}+\mathbf{A}_{ca}$ & $\mathcal{L}^{\ell}_{r}$ & $\mathcal{L}^{g}_{r}$ & Acc. \\ 
    \hline
    A &  &  &  & $\checkmark$ & $\checkmark$ & $\checkmark$ & 90.0 \\
    B & $\checkmark$ & & & $\checkmark$ & $\checkmark$ & $\checkmark$ & 90.6 \\
    C &              & $\checkmark$ &               & $\checkmark$                & $\checkmark$ & $\checkmark$ & 90.2 \\
    \hline
    D & $\checkmark$ & $\checkmark$ & $\checkmark$ & $\checkmark$ & $\checkmark$ & $\checkmark$ & 90.2 \\
    E & $\checkmark$ & $\checkmark$ & & & $\checkmark$ & $\checkmark$ & 90.4 \\
    \hline
    F & $\checkmark$ & $\checkmark$ & & $\checkmark$ & & & 90.0 \\
    G & $\checkmark$ & $\checkmark$ & & $\checkmark$ & $\checkmark$ & & 90.2 \\
    H & $\checkmark$ & $\checkmark$ & & $\checkmark$ & & $\checkmark$ & 90.6 \\
    \hline
    Ours & $\checkmark$ & $\checkmark$ & & $\checkmark$ & $\checkmark$ & $\checkmark$ & \textbf{91.0} \\
    \bottomrule
    \end{tabular}%
    }
    \caption{Module analysis of AIT and loss functions. $\mathbf{F}^{g*}$ means encoding $\mathbf{F}^{g}$ via Intra-frame Aligned Self-attention.} \label{tab:abla}
\end{table}

\paragraph{Effectiveness of Transformer Designs.}
We examine the effectiveness of our transformer design by conducting classification experiments under z/SO(3).
We first ablate one or both of the angular embeddings and report the results in Table~\ref{tab:abla} (models A, B, and C).
Model B performs better than model C by 0.4\%, which validates our design of feature integration where $\mathbf{M}^{\ell}_{i}$ is used as the main source of information.
When both angular embeddings are applied, the best result is achieved (\ie, 91.0\%).
Moreover, we validate our discussion in Section~\ref{sec:transformer} by comparing models D and E.
We demonstrate in model D that when encoding $\mathbf{F}^{g}$ in the same way as $\mathbf{F}^{\ell}$, the model performance decreases, which indicates that encoding $\mathbf{F}^{g}$ via self-attention will increase the model overfitting.
More analyses can be found in the supplementary material.
Finally, we examine the effectiveness of our attention logits-based integration scheme by comparing our model with the conventional method (model E), which applies self- and cross-attention sequentially and repeatedly.
We observe that our result is better than model E by 0.6\%, indicating that our design is more effective.

\paragraph{Registration Loss.}
We sequentially ablate $\mathcal{L}^{g}_{r}$ and $\mathcal{L}^{\ell}_{r}$ (models F, G, and H) to check the effectiveness of our registration loss deign.
Results in Table~\ref{tab:abla} demonstrate that we can still achieve a satisfactory result of 90.0\% without feature registration.
Individual application of $\mathcal{L}^{g}_{r}$ and $\mathcal{L}^{\ell}_{r}$ shows the improvement when forcing the final representation to be close to rotation-invariant features.
Moreover, it can be seen that model H performs better than model G, which indicates that intermediate features learned from the global scale are important for shape classification.
The best model performance is hence achieved by applying both losses.

\begin{figure}[t]
    \centering
    \includegraphics[width=\linewidth]{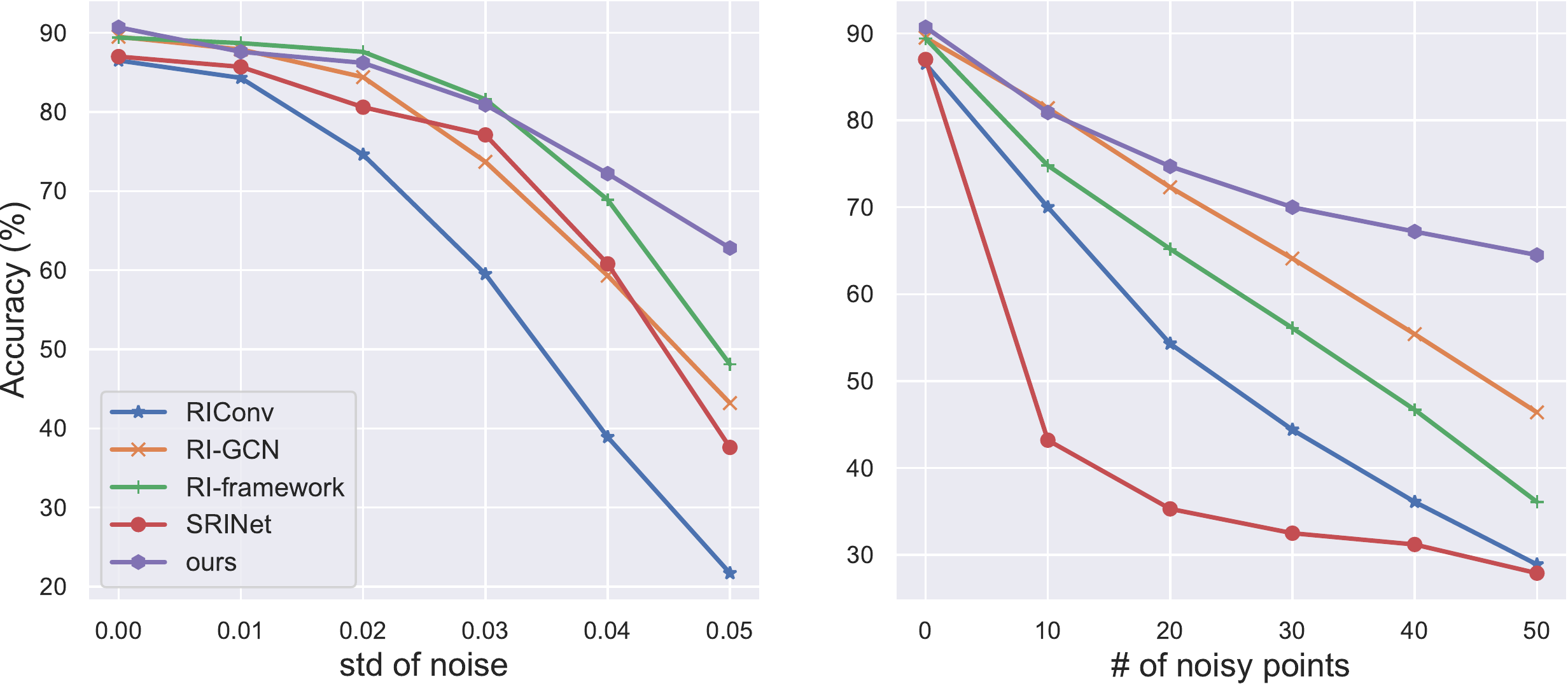}
    \caption{
    \textbf{Left}: Results on Gaussian noise of zero mean and variant standard deviation values.
    \textbf{Right}: Results on different numbers of noisy points.
    } \label{fig:noise}
\end{figure}

\paragraph{Noise Robustness.}
In real-world applications, raw point clouds contain noisy signals. 
We conduct experiments to present the model robustness to noise under z/SO(3).
Two experiments are conducted: (1) We sample and add Gaussian noise of zero mean and varying standard deviations $\mathcal{N}(0, \sigma^{2})$ to the input data;
(2) We add outliers sampled from a unit sphere to each object.
As shown in Fig.~\ref{fig:noise} (left), we achieve on par results to RI-Framework when std is low, while we perform better while std increases, indicating that our model is robust against high levels of noise.
Besides, as the number of noisy points increases, most methods are heavily affected while we can still achieve good results.

\begin{figure}[t]
    \centering
    \includegraphics[width=\linewidth]{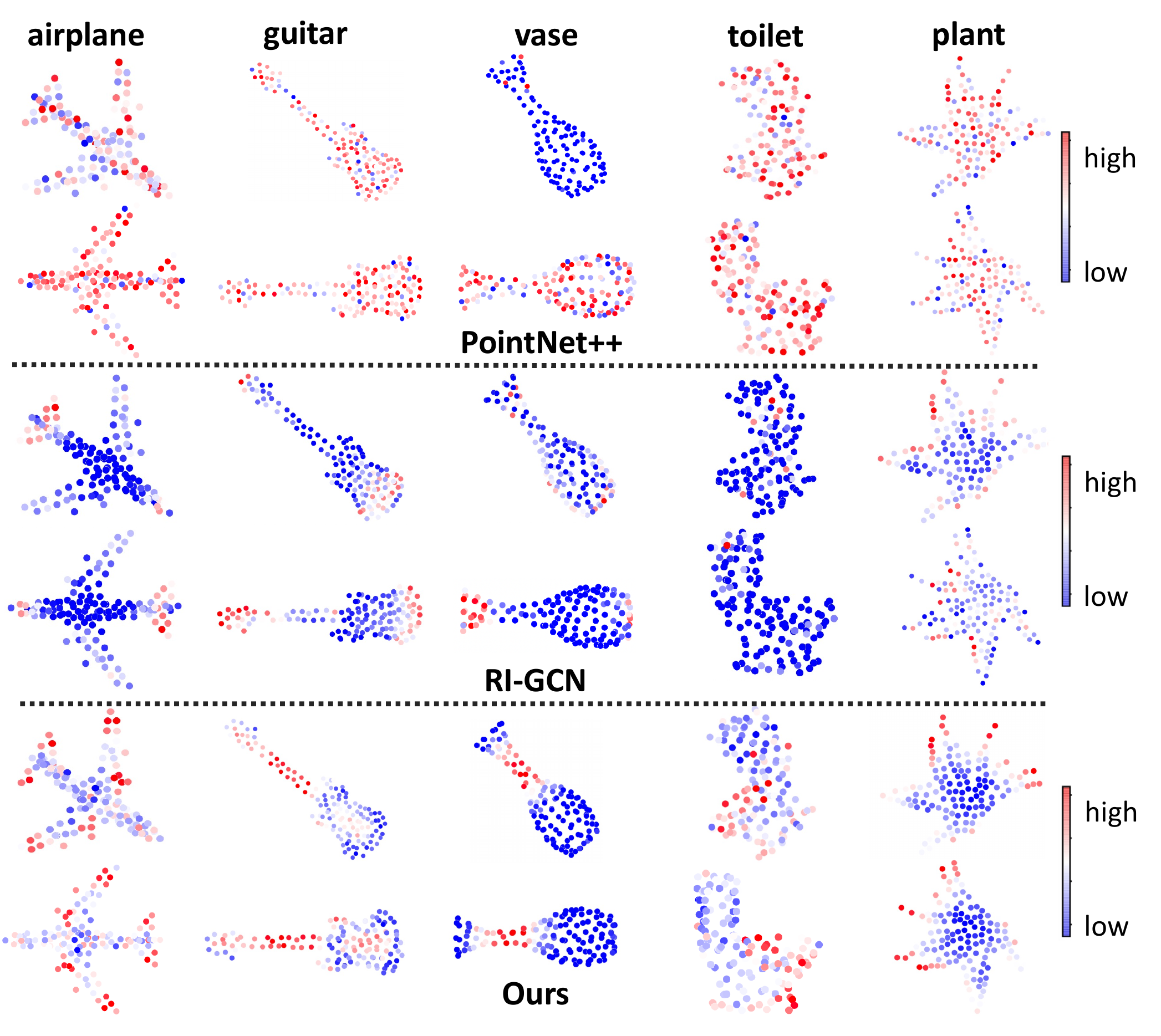}
    \caption{
    Network attention on PointNet++ (\textbf{top}), RI-GCN (\textbf{mid}) and our model (\textbf{bot}).
    } \label{fig:grad_cam}
\end{figure}
\paragraph{Visualization of Rotation Invariance.}
We further examine RI of learned features.
Specifically, we use Grad-CAM~\cite{grad_cam} to check how the model pays attention to different parts of data samples under different rotations.
Results are reported in Fig.~\ref{fig:grad_cam} with correspondence between gradients and colors shown on the right.
RI-GCN presents a good result, but its behavior is not consistent over some classes (\eg, vase and plant) and it does not pay attention to regions that are critical for classification (see toilet), showing inferior performance to ours.
PointNet++ shows no resistance to rotations, while our method exhibits a consistent gradient distribution over different parts with random rotations, indicating our network is not affected by rotations.

\section{Conclusion}
In this work, we rethink and investigate the close relation between rotation invariance and point cloud registration, based on which we propose a PCR-cored learning framework with three stages.
With a pair of rotation-invariant shape descriptors constructed from local and global scales, a comprehensive learning and feature integration module is proposed, Aligned Integration Transformer, to simultaneously effectively align and integrate shape codes via self- and cross-attentions.
To further preserve rotation invariance in the final feature representation, a registration loss is proposed to align it with intermediate features, where shared semantic knowledge of geometric parts is also extracted.
Extensive experiments demonstrated the superiority and robustness of our designs.
In future work, we will examine efficient methods for invariance learning on large-scale point clouds.
\newpage
\bibliography{ref.bib}
\end{document}


\maketitle
\section{Mathematical Proofs}
\paragraph{Proof of Rotation Equivariance of Orthonormal Basis $\mathbf{M}_{i}^{\ell}$ of LRFs.}
We prove the rotation equivariance of $\mathbf{M}_{i}^{\ell}$ designed for LRFs as mentioned in Section 3.1 of the main work.
Given a random rotation matrix $\mathbf{R} \in \mathbb{R}^{3 \times 3}$, it is easy to derive that $\vv{x_{i}}^{\ell}$ is equivariant to rotations given the rotated version $\vv{x_{i}}_{,rot}^{\ell}$:
\begin{equation}
\begin{aligned}
    \vv{x_{i}}_{,rot}^{\ell} & = \frac{\mathbf{R}p_i - \mathbf{R}p_m}{\|\mathbf{R}p_i - \mathbf{R}p_m\|_{2}} 
     = \frac{\mathbf{R}(p_i - p_m)}{\sqrt{\left(\mathbf{R}(p_i - p_m)\right)^{\top}\mathbf{R}(p_i - p_m)}} \\ 
    & = \frac{\mathbf{R}\vv{p_mp_i}}{\sqrt{\vv{p_mp_i}^{\top}\mathbf{R}^{\top}\mathbf{R}\vv{p_mp_i}}} = \mathbf{R}\frac{\vv{p_mp_i}}{\|\vv{p_mp_i}\|_{2}} 
    = \mathbf{R}\vv{x_{i}}^{\ell},
\end{aligned}
\end{equation}
where the subscript $rot$ represents the axis after rotations.
Moreover, $\mathbf{\Sigma}_{i}^{\ell}$ from Eq.~(1) of the main work after rotations can be represented as follows:
\begin{equation}
\begin{aligned}
    \mathbf{\Sigma}_{i, rot}^{\ell} & = \sum_{j=1}^{|\mathcal{N}(p_{i})|}\alpha_{j}\mathbf{R}\vv{p_ip_j}\vv{p_ip_j}^{\top}\mathbf{R}^{\top} \\ & = \mathbf{R}\left(\sum_{j=1}^{|\mathcal{N}(p_{i})|}\alpha_{j}\vv{p_ip_j}\vv{p_ip_j}^{\top}\right)\mathbf{R}^{\top} = \mathbf{R}\mathbf{\Sigma}_{i}^{\ell}\mathbf{R}^{\top}.
\end{aligned}
\end{equation}
As mentioned in the main work, eigenvalue decomposition can be directly applied to $\mathbf{\Sigma}_{i}^{\ell}$, resulting in the following expressions:
\begin{equation}
    \mathbf{R}\mathbf{\Sigma}_{i}^{\ell}\mathbf{R}^{\top} = \mathbf{R}\mathbf{U}_{i}^{\ell}\mathbf{\Lambda}_{i}^{\ell}{\mathbf{U}_{i}^{\ell}}^{\top}\mathbf{R}^{\top} = \left(\mathbf{R}\mathbf{U}_{i}^{\ell}\right)\mathbf{\Lambda}_{i}^{\ell}\left(\mathbf{R}\mathbf{U}_{i}^{\ell}\right)^{\top}.
\end{equation}
Since $\vv{z_{i}}^{\ell}$ is defined to have the same direction as the eigenvector with the smallest eigenvalue, after rotation, $\vv{z_{i}}_{,rot}^{\ell} = \mathbf{R}\vv{z_{i}}^{\ell}$.
Thus, the \textit{rotated} $y$-axis is:
\begin{equation}
\begin{aligned}
    \vv{y_{i}}_{,rot}^{\ell} 
    & = \vv{z_{i}}^{\ell}_{,rot} \times \vv{x_{i}}^{\ell}_{,rot} 
     = \mathbf{R}\vv{z_{i}}^{\ell} \times \mathbf{R}\vv{x_{i}}^{\ell} \\
    & = \operatorname{det}\left(\mathbf{R}\right)\left(\mathbf{R}^{-1}\right)^{\top}\left(\vv{z_{i}}^{\ell} \times \vv{x_{i}}^{\ell}\right) \\
    & = \mathbf{R}\left(\vv{z_{i}}^{\ell} \times \vv{x_{i}}^{\ell}\right)
    = \mathbf{R}\vv{y_{i}}^{\ell}.
\end{aligned}
\end{equation}
Since all basis vectors are rotation-equivariant, the local orthonormal basis $\mathbf{M}_{i}^{\ell} = [\vv{x_{i}}^{\ell}, \vv{y_{i}}^{\ell}, \vv{z_{i}}^{\ell}]$ is rotation-equivariant.

\paragraph{Proof of Rotation Invariance of Local Shape Descriptors $p_{ij}^{\ell}$.}
Here, we show the rotation invariance of local shape descriptors $p_{ij}^{\ell}$ introduced in Section 3.1 of the main work.
Based on the proof shown above, it is easy to show the rotation invariance of $p_{ij}^{\ell}$ after rotation $\mathbf{R}$:
\begin{equation}
\begin{aligned}
    p_{ij, rot}^{\ell}  &= \vv{p_ip_j}_{,rot}^{\top}\mathbf{M}_{i,rot}^{\ell} 
      = \left(\mathbf{R}\vv{p_ip_j}\right)^{\top}\left(\mathbf{R}\mathbf{M}_{i}^{\ell}\right)  \\
     &= \vv{p_ip_j}^{\top}\mathbf{M}_{i}^{\ell} = p_{ij}^{\ell}.
\end{aligned}
\end{equation}

\section{Model Details}
\begin{figure}
\centering
    \begin{subfigure}[b]{\columnwidth}
    \centering
    \includegraphics[width=\columnwidth]{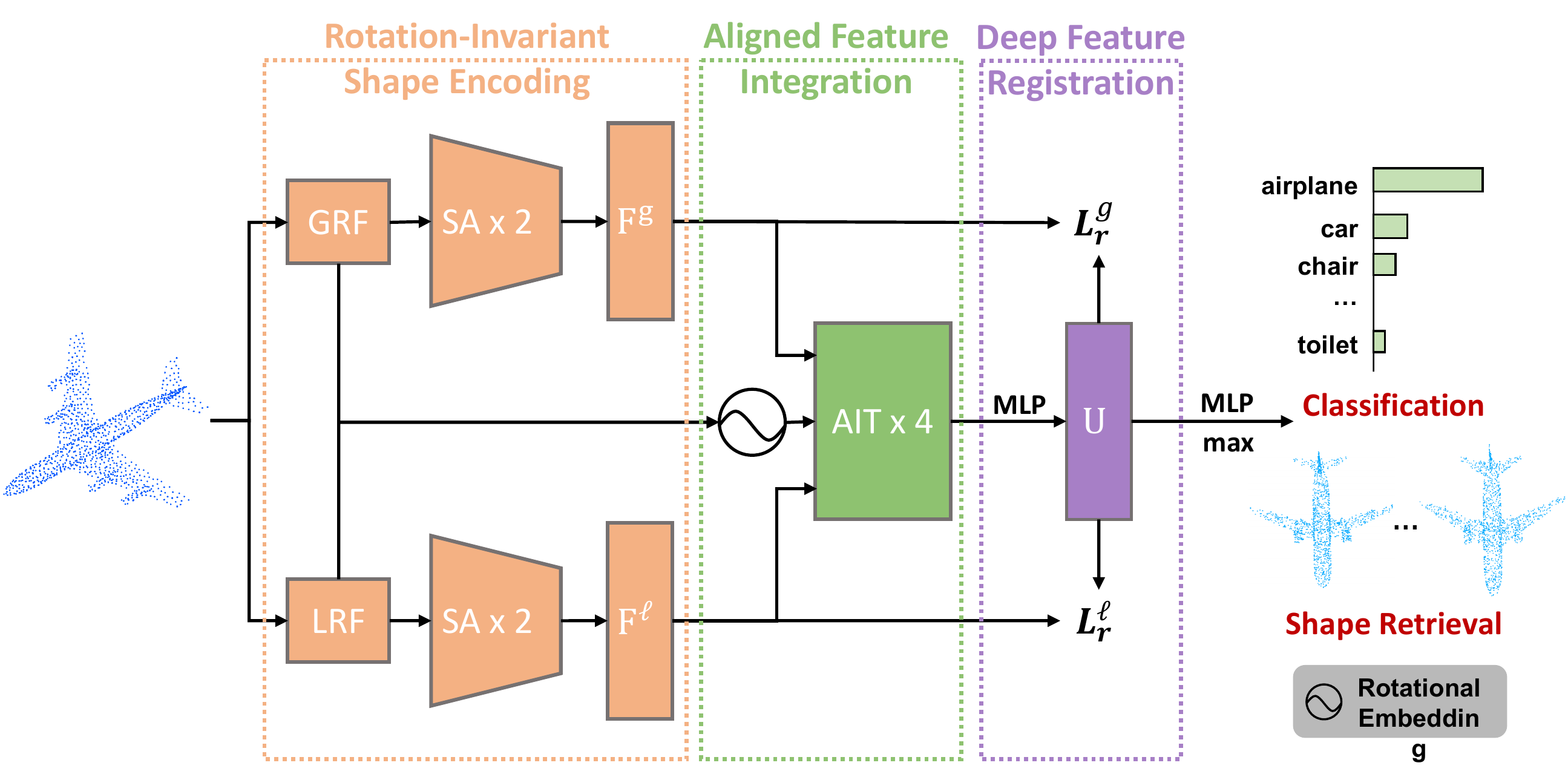}
    \caption{classification/retrieval}
    \label{fig:cls_model}
    \end{subfigure}
    \hfill
    \begin{subfigure}[b]{\columnwidth}
    \centering
    \includegraphics[width=\columnwidth]{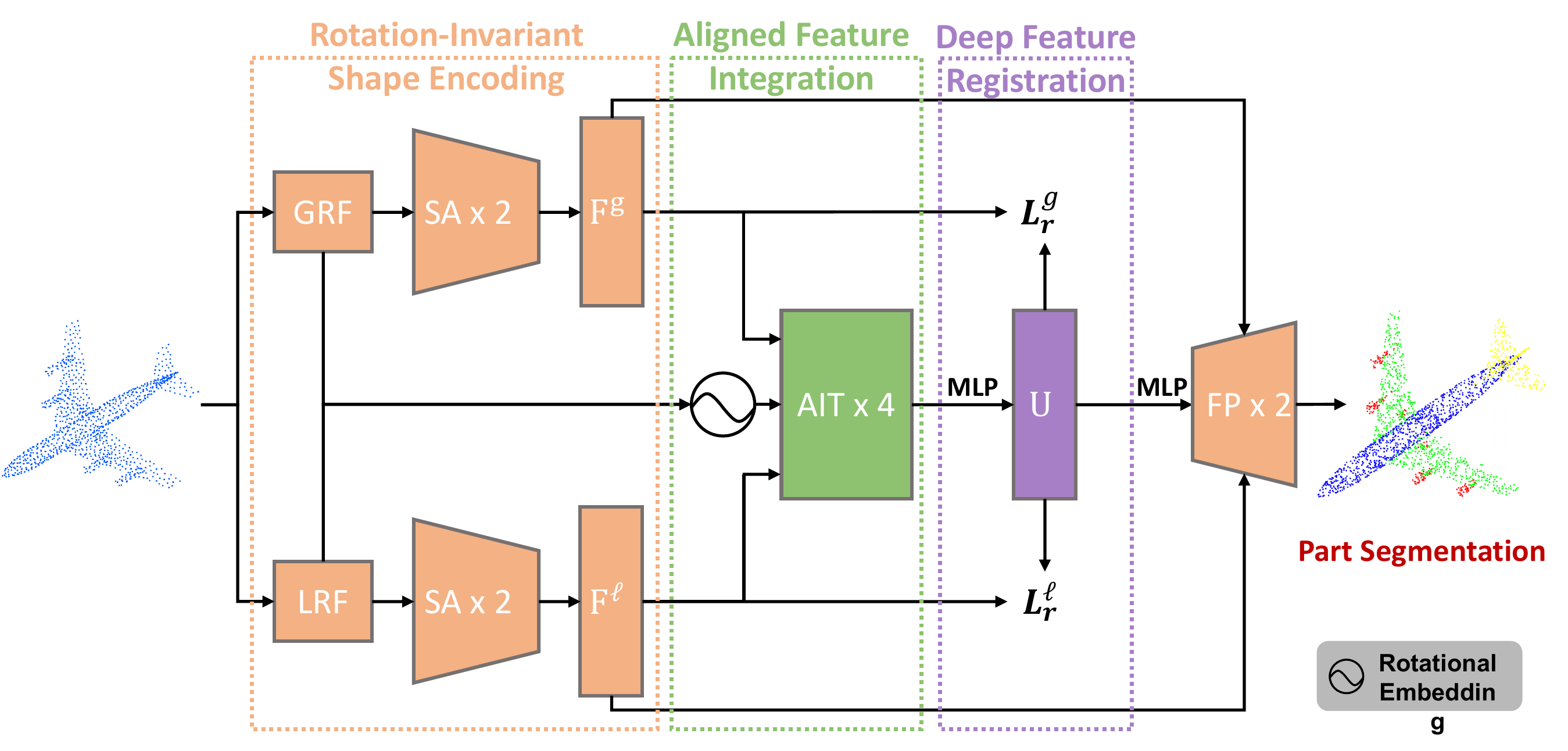}
    \caption{segmentation}
    \label{fig:seg_model}
    \end{subfigure}
\caption{
Model overviews for (a) classification / retrieval and (b) segmentation.
GRF: global reference frame; LRF: local reference frame; SA: set abstraction; AIT: Aligned Integration Transformer; and FP: forward passing.
}
\label{fig:model_overview}
\end{figure}

\begin{figure}
    \centering
    \includegraphics[width=0.85\linewidth]{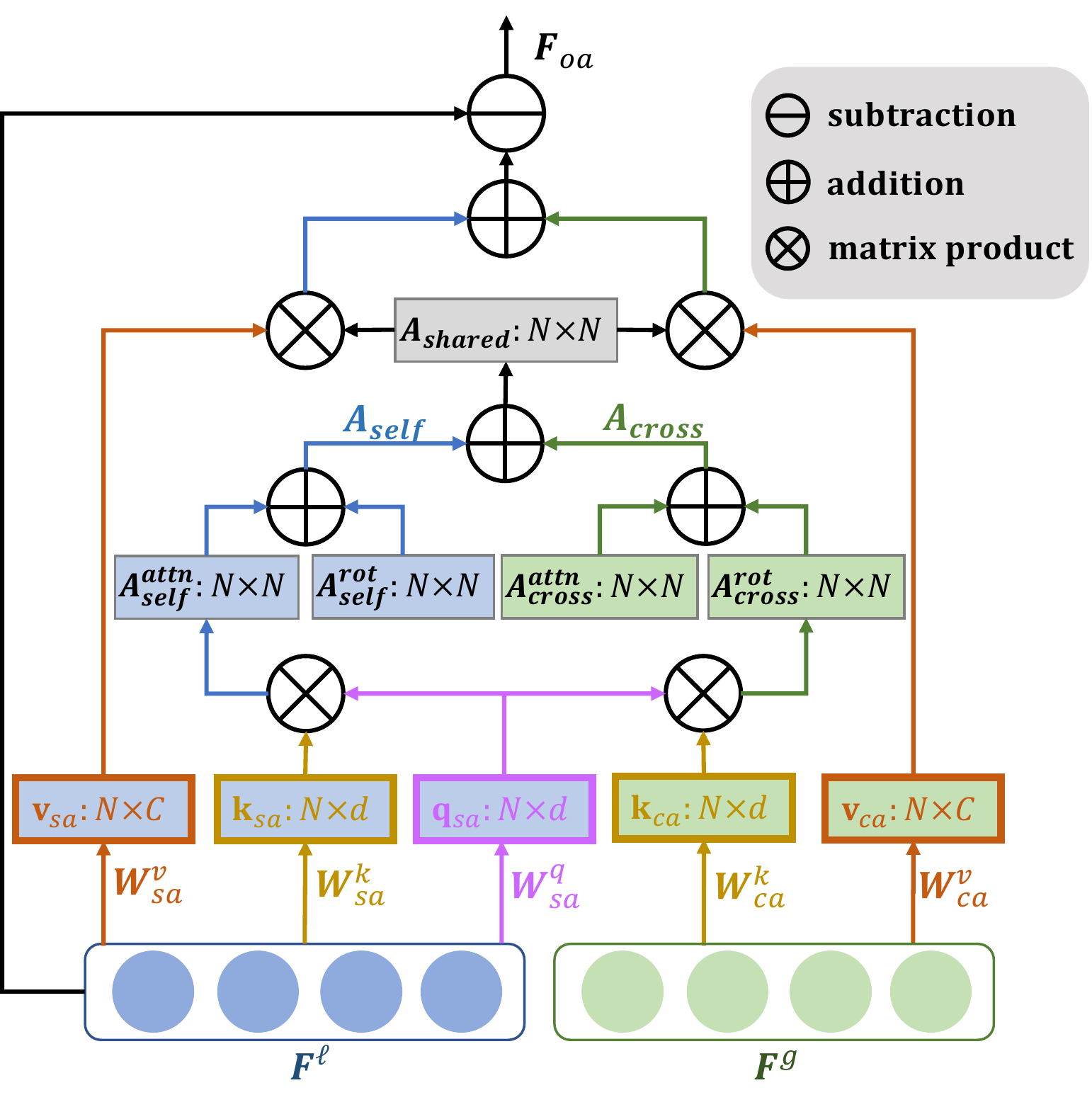}
    \caption{Illustrations of attention-based feature integration, where blue and green boxes indicate self- and cross-attention.
    \textcolor{Sienna}{Brown}, \textcolor{Goldenrod}{gold} and \textcolor{Purple}{purple} colored components correspond to $\mathbf{v}$, $\mathbf{k}$ and $\mathbf{q}$ implementations.}
    \label{fig:attention_fusion}
\end{figure}

\paragraph{Architectures.}
The model overview for the 3D classification, segmentation, and retrieval tasks is shown in Fig.~\ref{fig:model_overview}, where details of each module design are explained in Section 3 of the main work.
The architecture of attention-based feature integration module is illustrated in Fig.~\ref{fig:attention_fusion}.
Please refer to Section 3.2 of the main work for more details.

\paragraph{$\mathbf{F}^{g}$ in AIT.}
As mentioned in Section~3.2, to alleviate the model overfitting we do not apply self-attention on $\mathbf{F}^{g}$.
We argue that since we use global shape information as the supplementary material to assist local shape learning, this implementation allow lower-level information to flow effectively across layers to help the learning of higher-level local shape features, which could reduce the model overfitting.
We encode $\mathbf{F}^{\ell}$ in AIT blocks, as abstracting information from local structures can increase the model's ability on ﬁne-grained pattern recognition and generalizability to complex scenes~\cite{qi2017pointnet2}.
Hence, we find that without applying learnable modules on $\mathbf{F}^{g}$ is beneficial to our model performance.


\paragraph{Registration Loss.}
In this this part, we first explain the benefits of applying the registration loss to preserve the rotation invariance.
We then give more details about Eq.~(10) in the main work.

Suppose $\mathbf{U}^{\ell}$ and $\mathbf{U}^{g}$ are local and global part information of the final integrated feature $\mathbf{U}$. By maximizing the mutual information between $(\mathbf{U}, \mathbf{F}^{\ell})$ and $(\mathbf{U}, \mathbf{F}^{g})$, $(\mathbf{U}^{\ell}, \mathbf{F}^{\ell})$ and $(\mathbf{U}^{g}, \mathbf{F}^{g})$ are implicitly maximized. In this case, the shared geometric information between the local $\mathbf{F}^{\ell}$/global $\mathbf{F}^{g}$ and the integrated domain $\mathbf{U}$ are refined, increasing the representation power of $\mathbf{U}$.
Besides, the maximized similarities of $(\mathbf{U}^{\ell}, \mathbf{F}^{\ell})$ and $(\mathbf{U}^{g}, \mathbf{F}^{g})$ also tend to learn rotation invariance in an unsupervised manner.
Specifically, although $\mathbf{F}^{\ell}$/$\mathbf{U}^{\ell}$ encodes local patches with different poses (since LRFs are unaligned), and $\mathbf{F}^{g}$/$\mathbf{U}^{g}$ encodes the whole 3D object in a canonical pose, their feature similarity should be enforced to be similar as they represent the same 3D object, no matter what their poses are. Moreover, the mutual information between a local scale of a 3D object and a global scale of the same object is maximized, which embeds $\mathbf{U}$ with more accurate geometric information to distinguish it from objects in different classes.

We then explain the symbols in Eq.~(10).
$M$ stands for the set of all $B \times N$ pairs of positive samples across mini-batches, i.e., $M = \{(0, 0), ..., (B \times N, B \times N)\}$, where $B$ is the number of batch size and $N$ is the number of points.
The point feature $\mathbf{U}_{k}$ is the set of negative keys where  $(\cdot,k) \in M$ and $k \neq j$.
Note that since the registration loss function is applied to both $\mathbf{F}^{\ell}$ and $\mathbf{F}^{g}$, the point features of the same point encoded from the local and global scales are pushed close to each other, where the mutual information is increased such that shared geometric information can be discovered across the local and global scales.

\paragraph{Training Details.}
For all three tasks, we set the batch size to 32 for training and 16 for testing.
We use farthest point sampling to re-sample the points from the initial 10k points to 1024 points for classification and retrieval and 2048 points for segmentation.
Random point translation within $[-0.2, 0.2]$ and rescaling within $[0.67, 1.5]$ were adopted for augmentation.
We trained the model for 250 epochs with $t_{\alpha}$ = 15 and $t$ = 0.017.
SGD is adopted as the optimizer, where the learning rate was set to 1e-2 with momentum of 0.9 and weight decay of 1e-4.
Cosine annealing was applied to reschedule the learning rate for each epoch.
For classification and retrieval, we used one RTX2080Ti GPU with PyTorch for model implementation, and we used two GPUs for the segmentation task.
The normal vector information is ignored for all experiments.

\section{More Analysis Experiments}
\paragraph{Inﬂuence of Randomness.}
We report the variance and mean values of each model in Table 5 of the main work to derive a more accurate and reliable estimate of our model performance.
We hence report the variance and mean values of performance of each model in Table 5 on ModelNet40 with 5 training rounds.
As shown in Table~\ref{tab:random}, we can see that even for our model weights with the lowest performance 90.6\% (among our five repeated runs), it still surpasses the highest performance among models from A to H.

\begin{table}[!htb]
\centering
\resizebox{0.75\columnwidth}{!}{%
\begin{tabular}{l|ccc}
\toprule
Model     & A            & B            & C             \\
\hline
Acc. (\%) & 89.8$\pm$0.2 & 90.4$\pm$0.2 & 90.1$\pm$0.1   \\
\hline
Model     & D            & E             & F \\
\hline
Acc. (\%) & 89.8$\pm$0.4 & 90.1$\pm$ 0.3 & 89.6$\pm$0.4 \\
\hline
Model      & G            & H            & Best  \\
\hline
Acc. (\%)  & 90.0$\pm$0.2 & 90.3$\pm$0.3 & 90.8$\pm$0.2 \\
\bottomrule
\end{tabular}%
}
\caption{Variance and Mean values of different model performances on ModelNet40 with z/SO(3).}
\label{tab:random}
\end{table}

\paragraph{Point Re-sampling and Down-sampling.}
We examine different point re-sampling strategies from the initial 10k input points down to 1024 input points.
Experimental results of applying different sampling techniques on ModelNet40 are shown in Table~\ref{tab:resampling}, where we use random sampling (RS), farthest point sampling (FPS), uniform sampling (US), and inverse density importance sampling (IDIS) from \cite{groh2018flex} to examine the impact of different sampling methods on rotation invariance.
Note that point sampling affects both LRF and GRF constructions in our design, therefore we can only give analysis when considering both reference frames.
We can see that random sampling gives the lowest model performance with 89.7\%, with 1.3\% absolute performance drop compared to our method using FPS.
Inverse density importance sampling can achieve a comparable result as our method, while it is not strictly invariant to rotations.
We argue that due to the information compensation between features encoded from LRFs and GRF, different sampling strategies will not affect our model performance quite much.

\begin{table}[!htb]
\centering
\small
\begin{tabular}{l|cccc}
\toprule
Sampling Method & RS  & FPS (ours)  & US  & IDIS \\
\hline
z/z             & 89.7 & 91.0 & 90.2 & 90.6 \\
\hline
z/SO(3)         & 89.7 & 91.0 & 90.2 & 90.6 \\
\hline
SO(3)/SO(3)     & 89.7 & 91.0 & 90.2 & 90.6 \\
\bottomrule
\end{tabular}
\caption{Classification results (\%) on ModelNet40 with different re-sampling techniques.}
\label{tab:resampling}
\end{table}

\begin{figure}[!htb]
\centering
    \includegraphics[width=\linewidth]{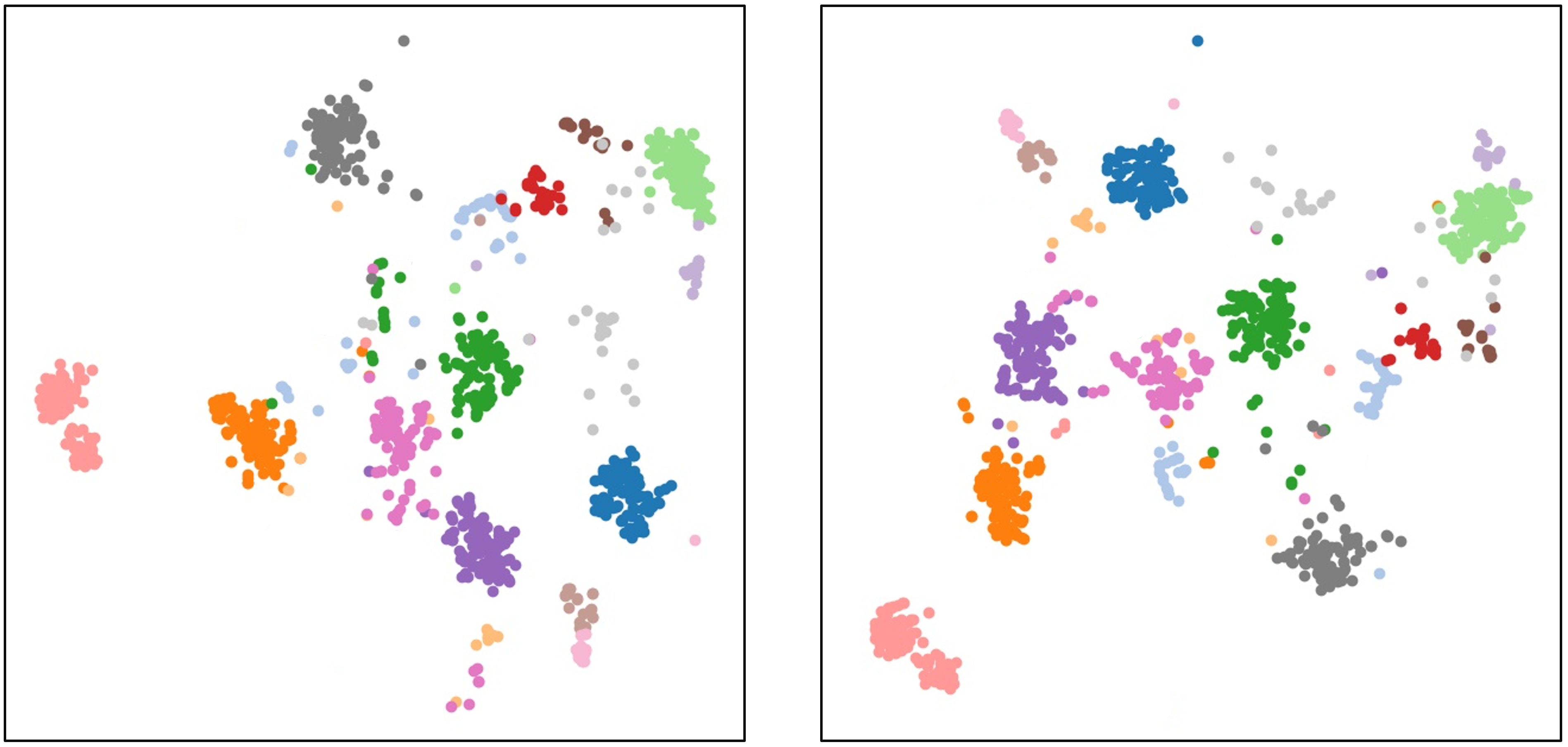}
    \caption{
    t-SNE of the aggregated $\mathbf{U}$ with z/SO(3) (\textbf{Left}) and SO(3)/SO(3) (\textbf{Right}).
    Clusters indicate good predictions in object classification.
    } \label{fig:sne}
\end{figure}

\paragraph{Visualization of $\mathbf{U}$.}
To better present the discriminability of the learned features, we summarize the shape feature representation $\mathbf{U}$ by maxpooling and visualize it via t-SNE~\cite{tSNE}.
Experiments are conducted on object classification under z/z and z/SO(3).
Only the first 16 classes are selected for a clear representation purpose as shown in Fig.~\ref{fig:sne}.
Although it is difficult to correctly separate all categories, we can see that some shape classes can be perfectly predicted, and the overall representation ability of $\mathbf{U}$ under different testing protocols is satisfactory and consistent.

\paragraph{Constructions of $p_{ij}^{\ell}$.}
We examine the model performance when using different methods to construct the local rotation-invariant feature $p_{ij}^{\ell}$.
Specifically, in addition to the proposed method that builds $p_{ij}^{\ell}$ based on LRFs, we examine point-pair features (PPFs) to build $p_{ij}^{\ell}$ following~\cite{ppfnet}.
\begin{table}[t]
\small
\centering
\begin{tabular}{l|cc}
\toprule
Method & PPFs & LRFs (Ours) \\
\hline
Acc. (\%) & 89.3 & 91.0 \\ 
\bottomrule
\end{tabular}
\caption{Classification results (\%) on ModelNet40 with z/SO(3).
}
\label{tab:lrf}
\end{table}
As reported in Table~\ref{tab:lrf}, we find that the model performance of using PPFs is lower than our LRF-based method.
The reason is that point positions, which provide information of exact shape of the 3D objects, are important for shape learning.
However, point-pair features give information about the topology of a 3D shape, and different 3D shapes can have the same topology, which introduces difficulties for exact 3D shape learning.

\begin{table}[b]
\resizebox{\columnwidth}{!}{%
\begin{tabular}{l|cccc}
\toprule
Method                                & Params (M) & Times (s) & Speed (ins./s) & Acc (\%) \\ \hline
RIConv                                & 0.68       & 0.041     & 396.4          & 86.4     \\ \hline
RI-GCN                                & 4.19       & 0.057     & 139.1          & 89.5     \\ \hline
RI-Framework                          & 2.36       & 0.134     & 43.1           & 89.4     \\ \hline
VN-DGCNN                              & 2.77       & -         & 77.3           & 89.5     \\ \hline
\citet{li2021closer}                  & 2.76       & 0.047     & 35.8            & 90.2    \\ \hline
Ours                                  & 3.11       & 0.043     & 205.3          & 91.0    
\\ \bottomrule
\end{tabular}%
}
\caption{Model complexity construction time for LRFs, and inference speed on ModelNet40 with z/SO(3), where \citet{li2021closer} is considered without test time augmentation.}
\label{tab:model_complex}
\end{table}
\paragraph{Model Complexity.}
Inference model sizes of different methods along with the corresponding construction time for LRFs and inference speed are reported in Table \ref{tab:model_complex}.
The construction time measured in seconds (s) shows time cost for different models generating their low-level rotation-invariant shape features, where we record the total time for local and global representation constructions of RI-Framework and our work.
VN-DGCNN does not compute the rotation-invariant shape features, therefore no result can be reported.
The inference speed with the unit of number of instances evaluated within one second (ins./s) is measured for each method with a batch size of 1.
When computing the inference speed, the amount of time for low-level rotation-invariant feature construction of methods~\cite{ri-framework, ri-gcn, RIconv, li2021closer} is also considered.
Table \ref{tab:model_complex} shows that our method only needs a relatively short construction time for both LRFs and GRF.
Meanwhile, the trade-off between the accuracy and inference speed is hard to balance.
We will investigate the model design for a much high accuracy and faster speeds in the future work.

\paragraph{Sign Ambiguity.}
As mentioned in the main work, we propose simple techniques to address the sign ambiguity issue introduced by eigenvalue decomposition when computing the LRFs and GRF.
We thus examine the model performance with no sign disambiguation techniques applied, of which the results are reported in Table~\ref{tab:ambiguity_cls}.
It can be seen that sign ambiguity negatively affects the model performance, where performances drop by 0.7\% and 0.9\% when uncertainty of vector directions is introduced to the model training.
With our proposed solutions, the model behavior can be stabilized hence the classification accuracy increases.

\begin{table}[htb!]
\centering
\small
\begin{tabular}{c|ccc} 
\toprule
Method    & no@$\mathbf{M}^{\ell}$ & no@$\mathbf{M}^{g}$ & no@$\mathbf{M}^{\ell}$ and $\mathbf{M}^{g}$ \\ 
\hline
Acc. (\%) & 90.3 & 90.1 & 89.8   \\
\bottomrule
\end{tabular}
\caption{Classification results (\%) on ModelNet40 with z/SO(3), where ``no@'' denotes no sign disambiguation technique applied.
}
\label{tab:ambiguity_cls}
\end{table}
\begin{table}[t]
\centering
\small
\begin{tabular}{l|cccc} 
\toprule
Method    & a & b & c & d (ours) \\ 
\hline
Acc. (\%) & 90.1 & 90.1 & 90.5 & 91.0 \\
\bottomrule
\end{tabular}
\caption{Classification results (\%) on ModelNet40 with z/SO(3).
}
\label{tab:valid_rotation}
\end{table}

\paragraph{Rotational Effect of $\mathbf{M}^{g}$.}
As mentioned in~\cite{li2021closer}, different ways to ensure $\mathbf{M}^{g}$ is a valid rotation matrix would result in different model performances.
In this part, we examine four different methods to ensure $\mathbf{M}^{g}$ is a valid rotation as follows:
(a) we randomly permute two basis vectors regardless of the $S$ value; (b) we randomly negate the value of a basis vector regardless of the $S$ value; (c) we permute two basis vectors of $S$ values being the smallest two; and (d) we simply reverse the direction of the basis vector whose S value is the smallest, which is the proposed method in our implementation.
We can see from Table~\ref{tab:valid_rotation} that our simple design achieves the highest value, while all the others decrease the model performance, which shows the effectiveness of our proposed method.

\begin{figure*}[t]
\centering\includegraphics[width=\linewidth]{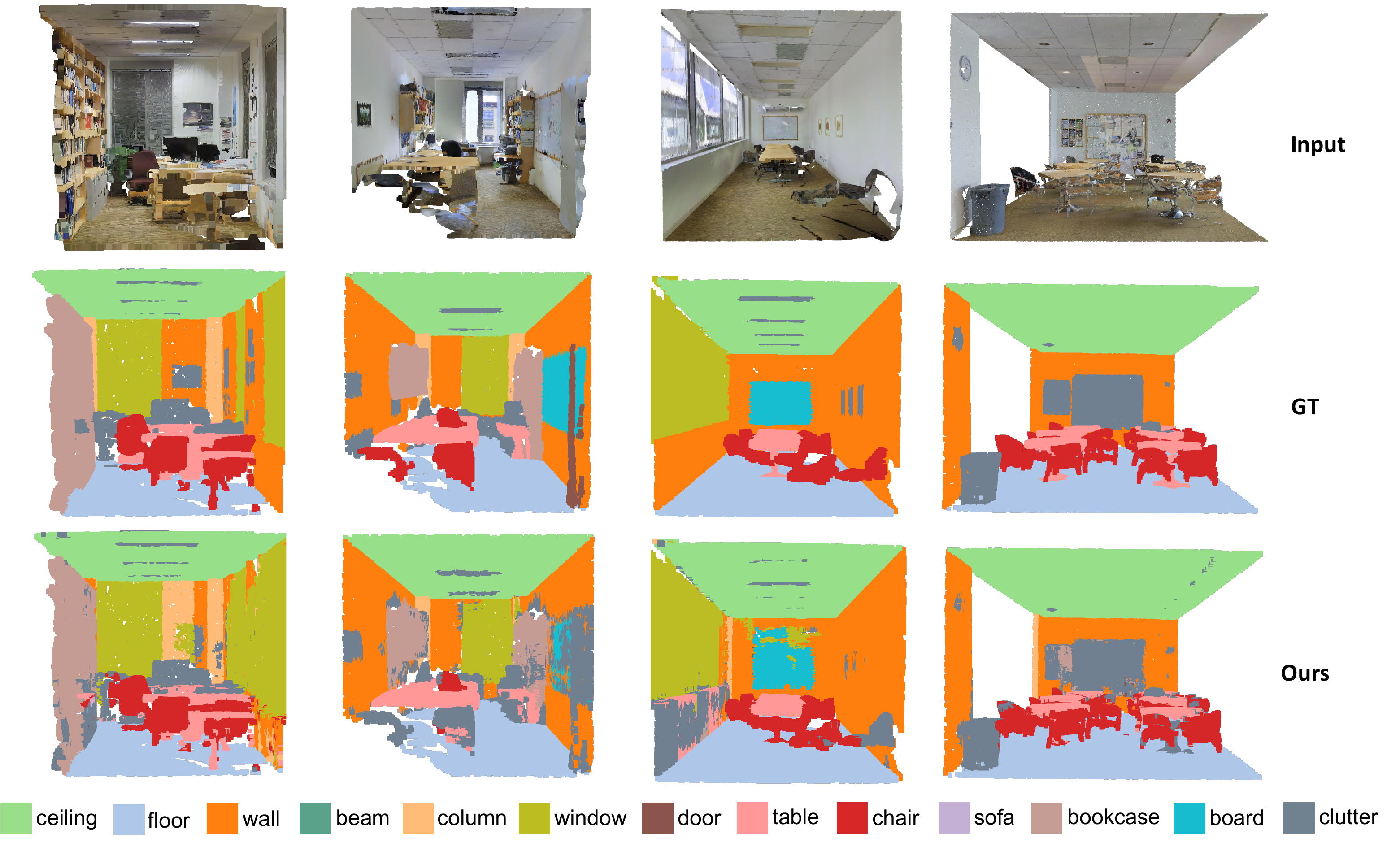}
\caption{
Visualization of semantic segmentation results on S3DIS area-5. The first row is the original inputs, the second row is the ground truth (GT) samples and the last row is our predicted results.
} \label{fig:s3dis}
\end{figure*}

\paragraph{3D Semantic Segmentation.}
To check our model's effectiveness on real-world large scenes, additional experiments are conducted on S3DIS dataset~\cite{s3dis}, which includes six indoor areas of three different buildings.
Each point is labeled by one of the 13 categories (e.g., ceiling, chair or clutter).
Following the same pre-processing steps as \cite{qi2017pointnet2, dgcnn}, each room is divided into 1m$\times$1m blocks and for each block 4096 points are sampled during training process.
We use area-5 for testing and all the other areas for training.
The quantitative results are shown in Table~\ref{tab:s3dis} following \cite{LGR-Net}, where it shows that under random rotations, our model outperforms LGR-Net by 7.8\%, showing a more effective way to process large indoor scenes. 
For a more intuitive understanding of our model performance, qualitative results are shown in Fig.~\ref{fig:s3dis} for reference.
\begin{table}
\centering
\resizebox{\columnwidth}{!}{%
\begin{tabular}{l|ccc} 
\toprule
Method    & z/z & z/SO(3) & SO(3)/SO(3) \\ 
\hline
PointNet \cite{qi2017pointnet} & 41.1 & 4.1 & 29.3 \\
DGCNN \cite{dgcnn} & 48.4 & 3.6 & 34.3 \\
\hline
RIConv \cite{RIconv} & 22.0 & 22.0 & 22.0 \\
LRG-Net \cite{LGR-Net} & 43.4 & 43.4 & 43.4 \\
Ours & \textbf{51.2} & \textbf{51.2} & \textbf{51.2} \\
\bottomrule
\end{tabular}%
}
\caption{
Semantic segmentation results (mIoU) on S3DIS area-5.
}
\label{tab:s3dis}
\end{table}

\begin{figure*}
\centering
\includegraphics[width=\linewidth]{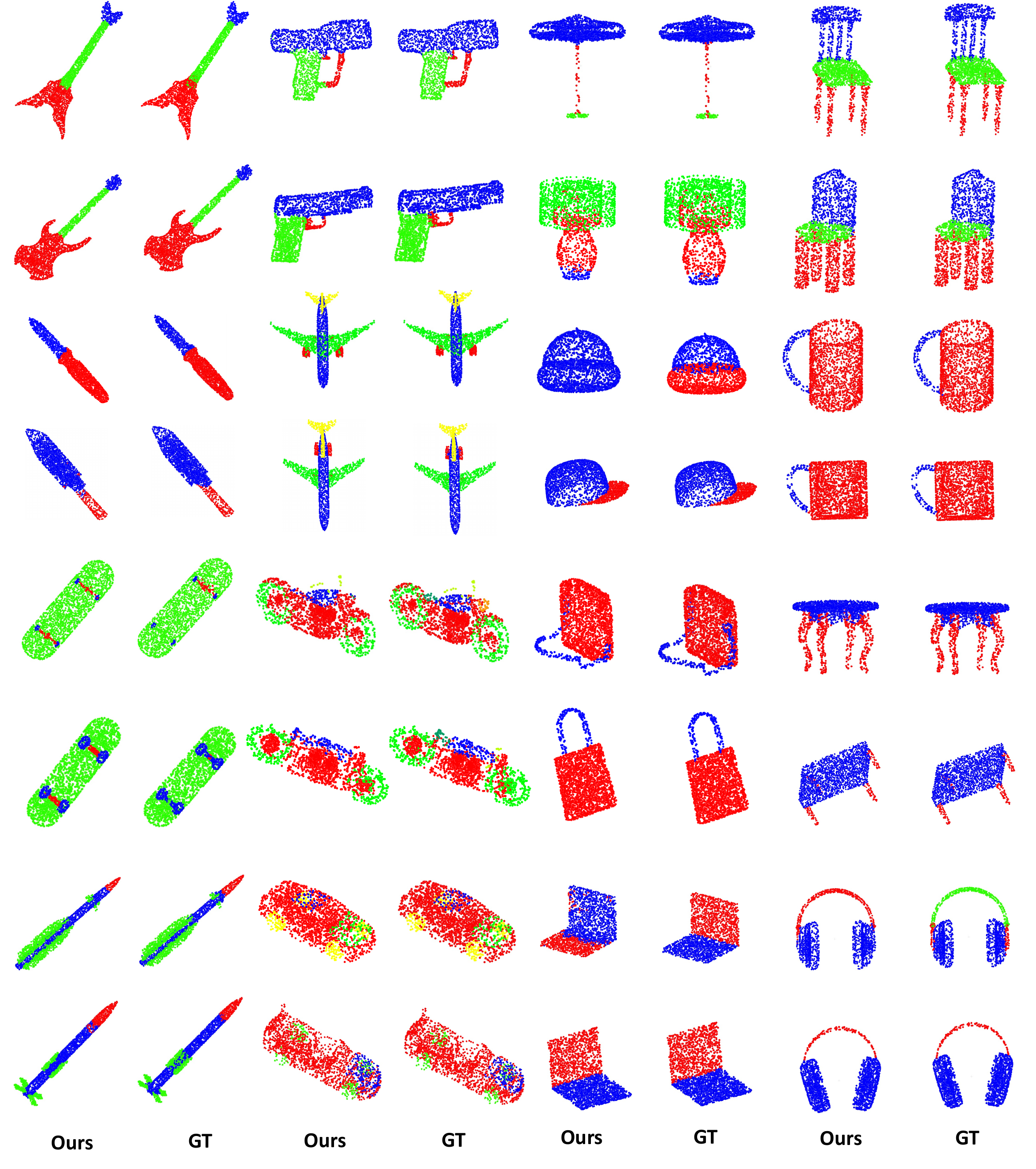}
\caption{Segmentation comparisons between the ground truth (GT) and our model on ShapeNetPart dataset under z/SO(3).
}
\label{fig:more_seg}
\end{figure*}

\paragraph{3D Part Segmentation.}
For visualization purposes (see Fig.~4 in the main work) as well as a detailed analysis of model behavior for each category, we report the per-class mIoU accuracies under z/SO(3) and SO(3)/SO(3) in Tables~\ref{tab:shapenet} and~\ref{tab:shapenet_so3}, where bold numbers indicate the best results for each category.
As per-class mIoU scores are not reported in VN-DGCNN~\cite{vnn}, we follow the official implementation\footnote[1]{https://github.com/FlyingGiraffe/vnn-pc} and report per-class mIoU results in both tables.
However, the reproduced results of VN-DGCNN are much lower than the ones reported in their work, and our model achieves better segmentation results than VN-DGCNN~\cite{vnn} for most categories.
Our model also outperforms the state-of-the-art methods~\cite{LGR-Net, luo2022equivariant} in several classes (\eg, airplane, chair, and table) under different testing conditions.
Furthermore, it is also obvious that our model performance is consistent across all categories when tested under different rotations.
In addition, we present more qualitative examples in Fig.~\ref{fig:more_seg}, which includes all 16 classes.
For each class, we show two pairs of ground truth and our predicted samples.
We can see that although errors occur when the boundary between the different parts have marginal difference, our model achieves great performance for most classes.

\begin{table*}[t]
\centering
\resizebox{\textwidth}{!}{%
\begin{tabular}{l|@{ }c@{ }|@{ }c@{ }@{ }c@{ }@{ }c@{ }@{ }c@{ }@{ }c@{ }@{ }c@{ }@{ }c@{ }@{ }c@{ }@{ }c@{ }@{ }c@{ }@{ }c@{ }@{ }c@{ }@{ }c@{ }@{ }c@{ }@{ }c@{ }@{ }c@{ }}
\toprule
\multirow{2}{*}{Method} & \multirow{2}{*}{mIoU} & air & \multirow{2}{*}{bag} & \multirow{2}{*}{cap} & \multirow{2}{*}{car} & \multirow{2}{*}{chair} & ear   & \multirow{2}{*}{guitar} & \multirow{2}{*}{knife} & \multirow{2}{*}{lamp} & \multirow{2}{*}{laptop} & motor & \multirow{2}{*}{mug} & \multirow{2}{*}{pistol} & \multirow{2}{*}{rocket} & skate & \multirow{2}{*}{table}  \\
& & plane & & & & & phone & & & & & bike & & & & board & \\ 
\hline
PointNet \cite{qi2017pointnet}           &37.8    &40.4    &48.1    &46.3    &24.5    &45.1    &39.4    &29.2   &42.6    &52.7    &36.7    &21.2    &55.0    &29.7    &26.6    &32.1    &35.8 \\
PointNet++ \cite{qi2017pointnet2}        &48.3    &51.3    &66.0    &50.8    &25.2    &66.7    &27.7    &29.7    &65.6    &59.7    &70.1    &17.2    &67.3    &49.9    &23.4    &43.8    &57.6 \\
PCT \cite{pct} &38.5    &32.2    &44.8    &36.3    &26.1    &36.2    &40.2    &48.1 &42.1    &54.0    &40.9    &18.7    &50.5    &25.6    &27.7    &44.7    &47.6 \\
\hline
RIConv \cite{RIconv} &75.3    &80.6    &80.0    &70.8    &68.8    &86.8    &70.3    &87.3   &84.7    &77.8    &80.6    &57.4    &91.2    &71.5    &52.3    &66.5    &78.4 \\
GCAConv \cite{GCA-Conv} &77.2    &80.9    &82.6    &81.0    &70.2    &88.4    &70.6    &87.1    &\textbf{87.2}    &81.8    &78.9    &58.7    &91.0    &77.9    &52.3    &66.8    &80.3 \\
RI-Framework \cite{ri-framework} &79.2    &81.4    &82.3    &\textbf{86.3}    &75.3    &88.5    &72.8    &90.3 &82.1    &81.3    &81.9    &\textbf{67.5}    &92.6    &75.5    &54.8    &75.1    &78.9 \\
LGR-Net \cite{LGR-Net}      &80.0 &81.5 &80.5 &81.4 &75.5 &87.4 &72.6 &88.7 &83.4 &83.1 &\textbf{86.8} &66.2 &92.9 &76.8 &62.9 &\textbf{80.0} &80.0
\\
VN-DGCNN$^{\star}$~\cite{vnn}         &75.3 &81.1 &74.8 &72.9 &73.8 &87.8 &55.9 &91.4 &83.8 &80.2 &84.4 &44.5 &92.8 &74.6 &57.2 &70.2 &78.9 \\
OrientedMP \cite{luo2022equivariant} &80.1 &81.7 &79.0 &85.0 &\textbf{78.1} &89.7 &\textbf{76.5} &\textbf{91.6} &85.9 &81.6 &82.1 &67.6 &\textbf{95.0} &79.6 &\textbf{64.4} &76.9 &80.7\\
\hline
Ours                        & \textbf{80.3} & \textbf{84.5} & \textbf{82.7} & 83.9 &76.6    &\textbf{90.2}    &76.1    &\textbf{91.6}    &86.6    &\textbf{83.5}    &84.6    &50.1    &94.4    &\textbf{81.9}    &60.3    &75.3    &\textbf{81.8} \\
\bottomrule
\end{tabular}%
}
\caption{Segmentation results of class-wise and averaged mIoU on ShapeNetPart under z/SO(3), where $\star$ means our reproduced results of VN-DGCNN using the official code.} \label{tab:shapenet}
\end{table*}

\begin{table*}[t]
\centering
\resizebox{\textwidth}{!}{%
\begin{tabular}{l|@{ }c@{ }|@{ }c@{ }@{ }c@{ }@{ }c@{ }@{ }c@{ }@{ }c@{ }@{ }c@{ }@{ }c@{ }@{ }c@{ }@{ }c@{ }@{ }c@{ }@{ }c@{ }@{ }c@{ }@{ }c@{ }@{ }c@{ }@{ }c@{ }@{ }c@{ }}
\toprule
\multirow{2}{*}{Method} & \multirow{2}{*}{mIoU} & air & \multirow{2}{*}{bag} & \multirow{2}{*}{cap} & \multirow{2}{*}{car} & \multirow{2}{*}{chair} & ear   & \multirow{2}{*}{guitar} & \multirow{2}{*}{knife} & \multirow{2}{*}{lamp} & \multirow{2}{*}{laptop} & motor & \multirow{2}{*}{mug} & \multirow{2}{*}{pistol} & \multirow{2}{*}{rocket} & skate & \multirow{2}{*}{table}  \\
& & plane & & & & & phone & & & & & bike & & & & board & \\ 
\hline
PointNet \cite{qi2017pointnet}           &74.4    &81.6 &68.7 &74.0 &70.3 &87.6 &68.5 &88.9 &80.0 &74.9 &83.6 &56.5 &77.6 &75.2 &53.9 &69.4 &79.9 \\
PointNet++ \cite{qi2017pointnet2}        &76.7    &79.5 &71.6 &87.7 &70.7 &88.8 &64.9 &88.8 &78.1 &79.2 &\textbf{94.9} &54.3 &92.0 &76.4 &50.3 &68.4 &81.0 \\
PCT \cite{dgcnn}                       &75.2 &80.1 &69.0 &82.5 &66.8 &88.4 &69.4 &90.4 &85.3 &81.8 &79.6 &39.9 &89.2 &76.5 &51.8 &72.6 &80.0 \\
\hline
RIConv \cite{RIconv}                    &75.5    &80.6 &80.2 &70.7 &68.8 &86.8 &70.4 &87.2 &84.3 &78.0 &80.1 &57.3 &91.2 &71.3 &52.1 &66.6 &78.5 \\
GCAConv \cite{GCA-Conv}                  &77.3 &81.2 &82.6 &81.6 &70.2 &88.6 &70.6 &86.2 &86.6 &81.6 &79.6 &58.9 &90.8 &76.8 &53.2 &67.2 &81.6 \\
RI-Framework \cite{ri-framework}         &79.4    &81.4 &\textbf{84.5} &\textbf{85.1} &75.0 &88.2 &72.4 &90.7 &84.4 &80.3 &84.0 &\textbf{68.8} &92.6 &76.1 &52.1 &74.1 &80.0 \\
LGR-Net~\cite{LGR-Net}                  &80.1 &81.7 &78.1 &82.5 &75.1 &87.6 &74.5 &89.4 &86.1 &83.0 &86.4 &65.3 &92.6 &75.2 &\textbf{64.1} &\textbf{79.8} &80.5 \\
VN-DGCNN$^{\star}$~\cite{vnn}        &74.7 &80.0 &79.4 &79.1 &71.5 &89.2 &66.1 &89.0 &83.5 &80.6 &82.0 &29.3 &91.4 &73.4 &51.5 &67.8 &81.0 \\
OrientedMP~\cite{luo2022equivariant}    &\textbf{80.9} &81.8 &78.8 &85.4 &\textbf{78.0} &89.6 &\textbf{76.7} &\textbf{91.6} &85.7 &81.7 &82.1 &67.6 & \textbf{95.0} &\textbf{79.1} &63.5 &76.5 &81.0 \\
\hline
Ours                                    &80.4    &\textbf{84.3} &82.2 &84.6    &77.9    &\textbf{89.9} &76.6  &91.3    &\textbf{86.7}    &\textbf{84.1}   &84.3    &50.1    &93.4  &79.0   &63.7 &75.3 &\textbf{82.3} \\
\bottomrule
\end{tabular}%
}
\caption{Segmentation results of class-wise and averaged mIoU on ShapeNetPart under SO(3)/SO(3).}
\label{tab:shapenet_so3}
\end{table*}

\bibliography{ref.bib}